\documentclass[10pt, runningheads]{llncs}

\usepackage{graphicx}
\usepackage{hyperref}

\usepackage{listings}                       

\usepackage[ruled]{algorithm2e}

\lstdefinestyle{our-style}{
    basicstyle=\scriptsize\ttfamily\linespread{0.8},
    captionpos=t,
    numbers=left,
    numbersep=5pt,
    keywordstyle=\color{blue},
    mathescape=true,
    breaklines=true,
    keywords={for, let, each, if, else, for, range, all, where}
}
\usepackage{chngcntr}

\usepackage{booktabs}                       
\usepackage{graphicx}
\usepackage{amsmath}
\usepackage{nicefrac}                       
\usepackage{svg}                            

\usepackage[inline]{enumitem}
\usepackage{tabularx}                       
\usepackage{bm}                             
\usepackage{todonotes}
\usepackage{float}                          
\usepackage{lipsum} 
\usepackage[font={small}]{caption}
\usepackage[font={small}]{subcaption}

\newcommand{\parent}[1][j]{\mathop{} p_{#1}}
\newcommand{\kde}[1][j]{\mathop{} \kappa_{#1}}

\newcommand{\instance}[1][i]{\mathop{} x_{#1}}

\newcommand{\iscore}[1][i]{\mathop{}\delta(\instance[#1])}

\newcommand{\cscores}[1][j]{\mathop{}\Lambda}
\newcommand{\iscores}[1][i]{\mathop{}\Delta}

\newcommand{\noisyor}[1]{\mathop{} \texttt{noisy-or}\big(#1\big)}

\newcommand{\dataset}{\mathop{} \mathcal{D}}

\newcommand{\attributes}{\mathop{} \mathcal{A}}

\newcommand{\squashparameter}{\mathop{} \rho}

\newcommand{\criticaldist}{$D_{crit}$}

\newcommand{\kul}{KU Leuven, Dept. of Computer Science, B-3000 Leuven, Belgium}
\newcommand{\leuvenai}{Leuven.AI - KU Leuven Institute for AI, B-3000 Leuven, Belgium}
\newcommand{\kulmail}{\texttt{\{firstname.lastname\}@kuleuven.be}}

\newcommand{\papertitle}{AD-MERCS:\\ Modeling Normality and Abnormality in Unsupervised Anomaly Detection}
\newcommand{\runningpapertitle}{AD-MERCS}

\newcommand{\admercs}{\texttt{AD-MERCS}}
\newcommand{\lof}{\texttt{LOF}}
\newcommand{\knn}{\texttt{kNN}}
\newcommand{\hics}{\texttt{HiCS}}
\newcommand{\hbos}{\texttt{HBOS}}
\newcommand{\iforest}{\texttt{iForest}}
\newcommand{\also}{\texttt{ALSO}}

\newcommand{\mercs}{\texttt{MERCS}}

\newcommand{\tb}[1]{\textbf{#1}}
\newcommand{\tu}[1]{\underline{#1}}

\title{\papertitle}
\titlerunning{\runningpapertitle}

\author{Jonas Soenen$^\star$, 
Elia Van Wolputte\thanks{Both authors contributed equally to the paper}, 
Vincent Vercruyssen, 
Wannes Meert, 
\and Hendrik Blockeel
}

\institute{\kul \\ \leuvenai \\ \email{\kulmail}}
    
\begin{document}
\maketitle              

\begin{abstract}
\small 
Most anomaly detection systems try to model normal behavior and assume anomalies deviate from it in diverse manners.
However, there may be patterns in the anomalies as well.
Ideally, an anomaly detection system can exploit patterns in both normal and anomalous behavior.
In this paper, we present \admercs{}, an unsupervised approach to anomaly detection that explicitly aims at doing both. 
\admercs{} identifies multiple subspaces of the instance space within which patterns exist, and identifies conditions (possibly in other subspaces) that characterize instances that deviate from these patterns.
Experiments show that this modeling of both normality and abnormality makes the anomaly detector performant on a wide range of types of anomalies. 
Moreover, by identifying patterns and conditions in (low-dimensional) subspaces, the anomaly detector can provide simple explanations of why something is considered an anomaly.
These explanations can be both negative (deviation from some pattern) as positive (meeting some condition that is typical for anomalies).
\end{abstract}

\section{Introduction}\label{s:intro}

Anomaly detection concerns the identification of ``abnormal'' instances in a data set. 
What is considered abnormal depends, of course, on the application context; there is no single definition of the term. 
An instance may be abnormal because it is a global outlier (i.e., it has values for certain attributes that deviate strongly from typical values), a local outlier (values are atypical among instances that are otherwise similar to this one), because it breaks a certain pattern in the data (consider e.g. the sequence ``\texttt{abababaababab}''), because it creates a pattern where none is expected (e.g. ``\texttt{vlajcoinjahjooooooookajdcna}''), and for many other reasons. 
As a result, many different approaches to anomaly detection exist, all with different strengths and weaknesses.

\begin{figure}[t]
    \centering
    \includegraphics[width=\textwidth]{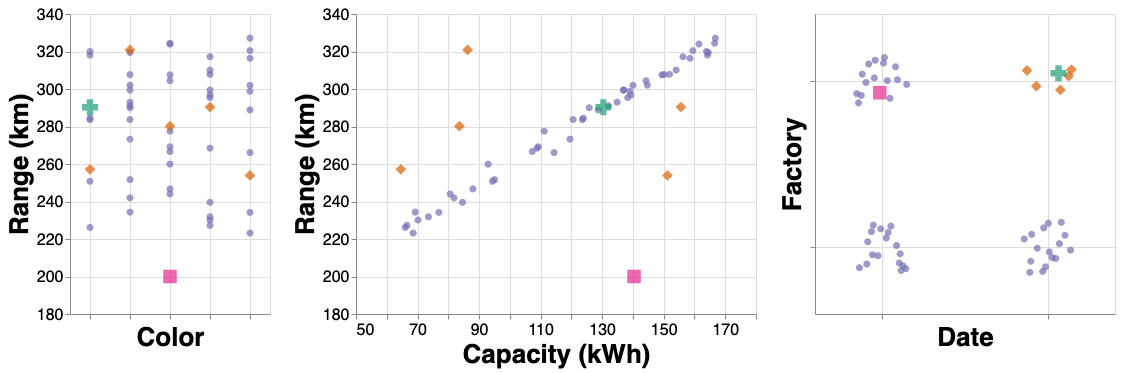}
    \caption{
    Electric vehicle (EV) dataset projected on three subspaces. The pink square is a global outlier: a $200$km range is never normal. Orange diamonds are local outliers: these EVs break the typical range-capacity pattern, but no individual range or capacity is unrealistic in isolation. The green cross is an accidental inlier: many other EVs made in this factory on this date are anomalous, raising suspicions about this particular EV.
    }
    \label{fig:illustration_guilty_by_association}
\end{figure}

In this paper, we investigate an entirely novel approach to anomaly detection that has a number of unique properties. 
The approach returns models that can both {\em identify} and {\em explain} anomalies, and these anomalies can be in the form of {\em local outliers}, {\em global outliers}, but also {\em accidental inliers}: instances for which there is reason to believe they are anomalies, even though they occur in a high-density region.
More specifically, the approach identifies low-dimensional subspaces in which certain patterns exist. 
An instance may be anomalous because it deviates from such a pattern, or -- and this is novel -- precisely because it adheres to the pattern. 
Figure \ref{fig:illustration_guilty_by_association} illustrates these different anomaly types on a toy dataset.

The new approach is based on so-called multi-directional ensembles of classification and regression trees (\mercs{})\cite{VanWolputte2018AAAI}. 
A \mercs{} model is an ensemble of decision trees, but differs from classical ensembles in that the trees it contains do not all try to predict the same target attribute, as in standard supervised learning settings.
Rather, any attribute can play the role of target variable in any given tree. 
A \mercs{} model thus contains a set of predictive trees, each of which expresses a pattern that governs the co-occurrence of values of any attributes. 
By the nature of decision trees, each such pattern typically involves relatively few attributes, and as such can be seen as identifying a low-dimensional subspace within which the discovered pattern is visible. 
By using standard tree-learning methods, the method identifies subspaces in which informative patterns are present.

Such a \mercs{} model could in principle be used to find anomalies by checking which instances violate the discovered patterns. 
The violated pattern can serve as an explanation of why the instance is seen as an anomaly. 
But because we have an ensemble of such patterns, which are to some extent independent views on what it means to be normal, more is possible. 
An instance can be considered more likely to be anomalous when it violates many different patterns. 
Now, when a single leaf of some decision tree contains many such instances, it is reasonable to assume that the leaf defines a set of conditions that are only fulfilled by anomalous instances; we call this an {\em anomalous context}. 
Instances in this leaf that were not yet (convincingly) identified as anomalies can now be assumed to be anomalous. 
Thus, an interaction is created between different subspaces, or ``views'' of the data, where two types of evidence can be exchanged and can reinforce each other: deviation from a normal pattern in one subspace, and belonging to an anomalous context in another subspace.

The above describes the basic intuition behind the proposed method, which is called \admercs{} (Anomaly Detection using \mercs{} models).  
Besides this, \admercs{} implements a number of other ideas, more on a technical level, which are explained in the technical sections of this paper.

In the remainder of this paper, we briefly describe the state of the art in anomaly detection, present the details of the novel \admercs{} approach, and present an empirical evaluation. 
The empirical evaluation is qualitative, showing that \admercs{} can indeed explain anomalies both in terms of normal patterns and in terms of anomalous contexts, and quantitative, showing that \admercs{} performs well over a wide range of anomaly detection problems. 
At the same time, the quantitative evaluation reveals some issues with current benchmarks that appear to have gone unnoticed till now.

\section{Related Work}\label{s:rel-work}
The detection of local and global outliers has been extensively studied in the literature.
A value for a given attribute is called a \textit{global outlier} if it falls outside of the spectrum of typical values.
It is called a \textit{local outlier} if it falls outside the spectrum of values typically observed \textit{among similar instances}. 
The set of similar instances is sometimes called a (local) context, and methods differ in how they define that context.

The above definition assumes a given attribute of interest. 
Often, there is no single attribute of interest, and an instance is called an anomaly as soon as it has one or more attributes with outlier values.
Identifying the attributes involved in the anomaly then becomes a task in itself. 
For local outliers, these attributes include not only the outlier attribute but also the attributes used for computing the local context.
We refer to this set of attributes as the relevant subspace.  

In what follows, we provide a brief overview of the anomaly detection literature and discuss how each approach quantifies the anomalousness of an instance, discovers contexts, and finds relevant subspaces.

\textbf{Nearest neighbor approaches} such as \knn{}\cite{dudani1976distance} and \lof{}\cite{Breunig2000SIGMOD} 
flag an instance as anomalous if it is far away from its nearest neighbors. \knn{} uses an absolute threshold for this, whereas \lof{} compares the distance to typical distances among other neighbors.
These methods implicitly identify relevant contexts (whatever is near constitutes the context) but not subspaces, as the distance metric predetermines the relevance of each dimension.

\textbf{Subspace-based methods} (\hbos{}, \iforest{} and \hics) detect anomalies in multiple subspaces and aggregate the results into a final anomaly score.
\hbos{}\cite{goldstein2012histogram} constructs a univariate histogram for each attribute.
An instance's anomaly score is the inverse product of the height of the bins it belongs to. Because \hbos{} uses univariate histograms, it detects global anomalies only.

\iforest{}\cite{Liu2008ICDM}, or \textit{isolation forest}, is an ensemble of \textit{isolation trees}, which isolate each instance in its own leaf by random splits.
\iforest{} assumes anomalies are easier to isolate and therefore have a shorter average path-length from root to leaf, across trees.
As the splits are random, \iforest{} uses highly randomized subspaces and contexts, and lacks interpretability. 

\hics{}\cite{Keller2012ICDE}, or \textit{High-Contrast Subspaces}, actively looks for subspaces $S$ where, for each attribute $A_i \in S$, the marginal $P(A_i)$ differs from its conditional distribution $P(A_i| S \setminus A_i)$.
Anomaly scores are then computed by running \lof{} in each subspace, and taking the average of those \lof{}-scores.
This combination benefits from \hics{}' subspaces and \lof{}'s contexts.

\textbf{Residual-based methods} such as \also{}\cite{Paulheim2015MLJ} convert the unsupervised AD-problem into a supervised learning problem.\footnote{Chandola et al.\cite{Chandola2009ACMCSUR} discuss this idea (under the moniker ``regression-based techniques'') mostly within time series AD, but it also applies to attribute-value data.}
\also{} learns for each attribute a model that predicts it from the other attributes. 
It computes an instance's anomaly score by calculating for each attribute the difference between its observed and predicted value (called a \textit{residual}), dividing that by the attribute's standard deviation, and aggregating this into one score.
Briefly summarized, an instance is flagged as anomalous if it violates many functional dependencies among attributes of normal cases.
Clearly, the choice of predictive model type (decision tree, k-NN, $\ldots$) determines the extent to which the anomaly detector identifies a context and a subspace, and to what extent it is interpretable.

Residual-based methods strongly rely on the assumption that normal instances are characterized by higher predictability of their attributes. However, it is perfectly possible that many attributes are inherently unpredictable (e.g., a coin flip), or that some attributes are more predictable among anomalies than among normal cases. 
As will become clear later on, residual-based methods can fail badly when their assumption is violated.

\textbf{Positioning of the proposed method.} 
Like residual-based methods, the proposed method \admercs{} finds relevant subspaces and contexts by learning a set of predictors.
However, \admercs{} explicitly avoids any supposed equivalence between predictability and normality.
First, to quantify the anomalousness of an instance, \admercs{} uses continuous density estimations instead of residuals.
This density-based mechanism enables \admercs{} to overcome a main drawback in residual-based methods: \admercs{} also captures ``unpredictable'' non-functional dependencies. 
Second, as mentioned in the introduction, \admercs{} recognizes that modeling abnormal behavior may help detect more anomalies, including ``accidental inliers'',  
such as the green cross in Fig.~\ref{fig:illustration_guilty_by_association}.
All AD methods discussed till now label individual instances: a conclusion about one instance does not affect the conclusion about another.
Identifying accidental inliers, however, requires recognizing commonalities among anomalies, meaning that conclusions about a group of instances will affect the conclusions about its individual members.
This is somewhat similar to the \textit{guilty-by-association} principle in graph-based anomaly detectors\cite{akoglu2015graph}, where nodes may be called anomalous just because they have many anomalous neighbors. The graph structure here provides a complementary view to the node labels. 
\admercs{} essentially uses the different subspaces created by the trees in the ensemble as complementary views to achieve a similar effect in the context of tabular data.

\section{AD-MERCS}\label{s:method}
Formally, \admercs{} tackles the following problem: 
given an $M$-dimensional dataset $\dataset=\{x_1, ..., x_N\}$ with $N$ instances $\instance= (x_i^1, \ldots, x_i^M)$, calculate an anomaly score $\iscore$ for each instance $\instance$.
Intuitively, $\delta$ should be higher for instances that in practice would more likely be considered anomalous. 
In this paper, that means instances that are deviating, or fulfill some condition that is typical for deviating instances.

To explain how \admercs{} works, we start from the basic \mercs{} approach, and then describe the adaptations made. 

\subsection{MERCS}\label{section:mercs}
Given a dataset $\dataset$ as described above, \mercs{} learns an ensemble of decision trees. 
For each attribute of the dataset, we learn one tree to predict that \textit{target} attribute.\footnote{Generally, in \mercs{}, a tree can predict multiple attributes and an attribute can be predicted by multiple trees. 
We do not use that functionality here.}
Trees are learned in standard, top-down fashion: attributes are selected for splits one at a time, based on their informativeness for the target attribute, and the resulting tree represents a function from a subset of the available input attributes to the target. 
Thus, with each tree $T_i$ corresponds a subspace $S_i$ that contains its input and target attributes, and this subspace is chosen so that the information in it is maximally predictive for the target; that is, a maximally strong ``pattern’’ is detected.
As a result, each leaf in a decision tree can be used as a context (a group of similar instances) for anomaly detection:
to decide whether a particular instance is anomalous, 
it can be compared with the instances with which it shares a leaf, as instances within a leaf are expected to adhere to the same pattern between input and target attributes.
We refer to~\cite{VanWolputte2018AAAI} for more details on \mercs{}.

Now, to do anomaly detection with a \mercs{} model we could use its decision trees in exactly the same way as \also{} uses its predictive models: aggregate the normalized residuals of all predictions made for a given instance.
However, this assumes that \textit{predictability} and \textit{normality} are equivalent, and we argued earlier that this assumption is problematic in two ways.
First, unpredictable yet normal behavior (e.g. a coin flip) is wrongly seen as anomalous.
Second, predictable yet abnormal behavior escapes detection, as any predictable behavior leads to low residuals. 
The contributions of this work are essentially solutions to these two issues, and in the remainder of this section we explain how they work:
in Section~\ref{section:mercs_with_density}, we introduce a density-based scoring mechanism so \admercs{} can handle unpredictable behavior; 
in Section~\ref{section:admercs}, we ensure that \admercs{} recognizes contexts where abnormal behavior is the norm as anomalous contexts.

\subsection{MERCS with Density Estimation}\label{section:mercs_with_density}
\begin{figure}[t]
    \centering
    \includegraphics[width =\textwidth]{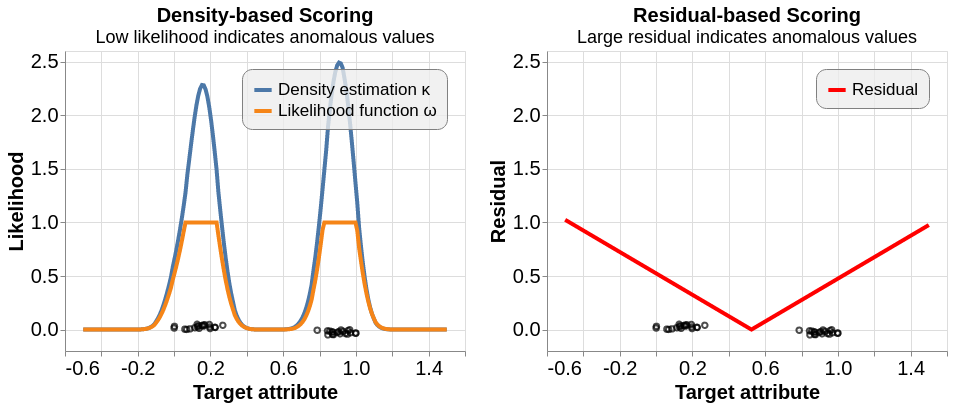}
    \caption{
    Density-based and residual-based scoring for a leaf where target values (black circles) follow a bimodal distribution. \textbf{Left.} Density-based scoring correctly considers $0.5$ as anomalous (likelihood of 0). The density estimation $\kappa$ is shown in blue and the likelihood function $\omega$ used for scoring is in orange. 
    \textbf{Right.} Because the tree predicts $0.5$ for any instance classified in this leaf, residual-based scoring considers instances with a target value of $0.5$ perfectly normal in this leaf (the residual is $0$), while this value can reasonably be assumed anomalous as it has never been observed before. }
    \label{fig:residual_vs_density}
\end{figure}

The first contribution of this work introduces a density-based scoring mechanism in \admercs{}.
As explained before, residual-based methods do not work well under all circumstances. 
For instance, suppose a leaf contains instances with target values $1.52, 1.55, 1.57, 2.81$. The value 2.81 is an outlier, and indeed, if the mean of the leaf is used as a prediction, we will find that 2.81 has a greater residual than the other values.  
In contrast, consider a leaf with values $1.4, 1.5, 1.6, 3.4, 3.5, 3.6$. 
The mean value for this leaf is 2.5, but there are clearly two clusters here.  
If an instance with value 2.5 gets sorted into this leaf, it can reasonably be considered an outlier (it does not fit the clusters), yet it has a zero residual.
Clearly, the basic idea behind residual-based predictions does not work well here (Fig.~\ref{fig:residual_vs_density}). 

In \admercs, instead of using residuals, each tree in the ensemble scores the instances based on the likelihood of its target value as indicated by a histogram (if the target is nominal) or some local probability density (if the target is numeric). 
More specifically, in each leaf, \admercs{} estimates the density $\kappa$ of the target attribute  using Gaussian kernel density estimation\cite{scott2015}, selecting the kernel bandwidth using the approach of Botev et al.~\cite{botev2010}.

Inside a leaf, lower values of $\kappa$ indicate less likely target attribute values.
However, there are two reasons we cannot do anomaly detection using $\kappa$ values directly.
First, distinguishing high values of $\kappa$ is pointless: all those indicate ``normality'' anyway.
Second, $\kappa$ values are incomparable between leaves: densities are normalized in terms of their area, not height, so their values depend on the local range of the target variable. 
As a solution, the $\kappa$ values are translated into likelihood values $\omega$ using the following equation:
\begin{equation}
\label{e:squashing}
\omega_j(x) =  
        \begin{cases}
            \frac{\kde(x)}{\tau_j(\squashparameter)} \quad &\kde(x)< \tau_j(\squashparameter) \\
            1 \quad &\kde(x)\geq \tau_j(\squashparameter)
        \end{cases}
\end{equation}
where threshold $\tau_j(\rho)$ is exceeded by $\rho\%$ of the instances in leaf $j$. In other words, the  $(100-\rho)\%$ lowest $\kappa$ values are linearly mapped to the interval $[0,1]$, the rest is mapped to $1$.
The closer to $0$ an instance’s $\omega$ value is, the more likely it is anomalous. 
Fig.~\ref{fig:residual_vs_density} illustrates this density-based scoring on a bimodal distribution of target values.

So far, we assumed this likelihood estimation to take place in each leaf; 
however, a more thorough analysis shows room for further improvement.
We use decision trees to find low-impurity leafs, but after a split, the impurity of a child node may still exceed the impurity of its parent. 
This happens if the impurity \textit{decrease} from the parent to one child node is large enough to compensate for the impurity \textit{increase} from the parent to the other child node.
The pattern that holds in the parent is then actually stronger than the pattern that holds in the high-impurity child node.
Therefore, after learning the full tree, if a leaf $l_j$ has a higher impurity than one of its ancestors, the instances classified in that leaf are scored using the likelihood function of the lowest-impurity ancestor of $l_j$. 
This ensures that we always use the best pattern available to score a particular instance. 

Using the procedure above, each tree in the ensemble assigns its own anomaly score to each individual instance. 
To aggregate these scores, we interpret the anomaly scores $1 - \omega_j(x)$ as probabilities $p_i$ and aggregate them using a \mbox{noisy-\textsc{or}} gate\cite{galan2000modeling} with inhibition probability $1-\gamma$ for each input: 
$$ 
\texttt{noisy-or}\left({ \{p_i\}_{i=1, \ldots, n}; \gamma}\right) = 1 - \prod_i (1- \gamma p_i).
$$ 
Intuitively, using an \textsc{or} gate reflects that an instance is considered an anomaly if it looks anomalous in at least one subspace.
Inhibition reduces the trust we put in every single estimate: a higher inhibition probability (a lower $\gamma$) causes the aggregation to require more evidence, from all subspaces combined, before an instance gets assigned a high anomaly score.

This procedure yields an anomaly detector that can model non-functional dependencies among normal cases, but does not yet explicitly model anomalies.
In the following subsection, we describe how \admercs{} detects anomalous contexts (i.e. contexts where abnormal behavior is the norm) and how this influences the anomaly scores. 

\subsection{AD-MERCS}\label{section:admercs}
To address the flawed assumption that abnormal behavior is necessarily unpredictable, the second contribution of this work ensures that \admercs{} can identify anomalous contexts. By doing so, \admercs{} can also exploit patterns in abnormal behavior, and detect more anomalies.

Assume for a moment that we know which instances are anomalous. 
If a leaf exclusively contains anomalous instances, it is reasonable to assume that the conditions defining the leaf define an anomalous context;
in fact, this is standard in supervised anomaly detection where new instances are considered anomalous if they end up in such a leaf.
We want \admercs{} to have a similar capacity to flag an instance as anomaly simply because it belongs to an anomalous context.

This is not trivial because \admercs{} works in an unsupervised manner, but the fact that a single \admercs{} model contains many trees makes an iterative approach possible. 
In the first iteration, we compute an anomaly score $\delta_i$ for each training instance $x_i$, using the approach explained above. 
Based on these scores, we can now assign an anomaly value to each context (i.e. leaf) of each tree: a context with many high-scoring instances is more likely to be an anomalous context. 
Next, these context scores can be used to adapt the anomaly scores of individual instances: when an instance appears abnormal in a normal context \textit{or} belongs to an anomalous context, its score is raised.  
On the basis of the new scores, context scores are recomputed, and so on, until convergence or a stopping criterion is reached.
Note that the whole procedure does not require retraining the ensemble, and the tree structures remained unchanged during the process.  

Specifically, after obtaining the $\omega_j$ functions that indicate how anomalous any instance is in context $c_j$, \admercs{} updates an array of $\delta_i$ values (one per instance $x_i$ in the training set) and $\lambda_j$ values (one per context $c_j$), using the following update rules:
\begin{align*}
v_{ij} = \lambda_j + (1-\lambda_j) (1-\omega_j(x_i)) \label{eq:v_ij}\\
\delta_i = \noisyor{ \{v_{ij} | x_i \in c_j\}; \gamma_\delta} \\
\lambda_j = 1-\noisyor{\{1-\delta_i | x_i \in c_j)\}; \gamma_\lambda}
\end{align*}

The $v_{ij}$ values are context-specific estimates for the anomalousness of an instance $x_i$ that take into account both the (current estimate of the) abnormality of context $c_j$, and its own abnormality in that context. 
They are essentially a weighted mean of 1 (if the context is anomalous) and $1-\omega_j(x_i)$ (if it is normal).
These estimates are aggregated into a global estimate $\delta_i$ for the abnormality of $x_i$, as explained before. 
The $\delta_i$ values of all the examples in a context are next aggregated into a single $\lambda_j$ value for that context, using what is essentially a noisy-\textsc{and}: a context is considered anomalous when it covers only anomalous instances. 
The $\gamma_\delta$ and $\gamma_\lambda$ parameters provide some control over how strong the combined evidence should be.

Algorithm~\ref{a:algo} summarizes the \admercs{} method in its entirety.  It consists of three phases.
First, all trees are learned and their leaves are returned as contexts;
next, for each leaf the density is estimated (using the leaf itself or its ancestor with the lowest impurity); 
finally, the interleaved computation of instance scores and context scores is performed.

\begin{algorithm}[t]
\caption{\admercs{}}
\label{a:algo}
\SetKwInOut{KwIn}{In}
\SetKwInOut{KwOut}{Out}
\KwIn{\small Dataset $\dataset$ with attributes $\attributes$, normalization constant $\squashparameter$, inhibition probabilities $\gamma_\delta$ and $\gamma_\lambda$, number of iterations $n$}
\KwOut{\small Instance anomaly scores $\boldsymbol{\delta}$, context anomaly scores $\boldsymbol{\lambda}$}
\begin{lstlisting}[style=our-style]
# phase 1: discover contexts
for $i=1, \ldots, |\attributes|$:
    $T_i$ = learn_tree($\dataset$, inputs=$\attributes \setminus \{A_i\}$, outputs=$\{A_i\}$)
let $l_j$ denote leaf $j$ (numbered consecutively throughout ensemble)
let $c_j$ denote the set of instances covered by $l_j$

# phase 2: model typical behavior
for each leaf $l_j$:
    $\parent =\texttt{get\_ancestor\_with\_lowest\_impurity}(l_j)$
    if impurity($l_j$) > impurity($\parent$):
        $\kde = \texttt{do\_kde}(p_j)$
    else:
        $\kde = \texttt{do\_kde}(l_j)$
    $\omega_j$ = normalize_kde($\kde, \squashparameter$) # function $\omega_j$ from Eq. 1

# phase 3: estimate instance and leaf abnormalities
$\boldsymbol{\lambda} = \boldsymbol{0}$ # Initially, every context is considered normal
for $\_$ in range($n$):
    for all $i$, $j$ where $x_i\in c_j$: $v_{ij} = \lambda_j + (1-\lambda_j) (1-\omega_j({x_i}))$
    for all $i$: $\delta_i = \noisyor{ \{v_{ij} | x_i \in c_j\}; \gamma_\delta}$
    for all $j$: $\lambda_j = 1-\noisyor{\{1-\delta_i | x_i \in c_j)\}; \gamma_\lambda}$
\end{lstlisting}
\end{algorithm}

\subsection{Explanations}
Finally, \admercs{} does not only detect anomalies but also explains them on a level that domain experts understand.
For each instance $x_i$ flagged as anomalous, \admercs{} knows exactly which trees were responsible and offers concise explanations of how these trees arrived at this conclusion.
Similarly, for each anomalous context, \admercs{} immediately provides an interpretable description of this group as a whole.
This greatly facilitates the extraction of actionable knowledge from the AD-process.

Concretely, if a tree $T$ assigns an individual instance $x_i$ a high $v_{ij}$ value, that instance is deemed anomalous by that tree.
Note that this can be due to either $x_i$'s abnormal behavior within a normal context, or simply because it belongs to an anomalous context.
We discuss these two scenarios separately.

First, if $x_i$ belongs to a \textit{normal} context $c_j$, it apparently violates the pattern that holds in the node $n$ used to score instances in $c_j$. 
In this case, the explanation consists of two parts: an interpretable description of node $n$, which can be obtained by traversing the tree; and the normal target values, as captured by the likelihood function $\omega_j$, together with instance $x_i$'s atypical target value.

Second, if $x_i$ belongs to an anomalous context $c_j$, all we really need as explanation is an interpretable description of this context: in this scenario, the context description characterizes the anomalies directly. 

\subsubsection{Illustration}
We illustrate \admercs{}' explanations using the Zoo dataset\cite{Dua:2019}, which describes different animals. 
The most anomalous animals according to \admercs{} are scorpion, platypus and seasnake. 
Fig.~\ref{fig:zoo_illustration} shows how \admercs{} explains why the scorpion is anomalous.
Several reasons are mentioned. 
Perhaps the most appealing one is that scorpions, being invertebrates, have a tail (invertebrates typically do not have a tail).  Not having a backbone defines the context here, ``tail'' is the attribute with the atypical value. 
Further, animals that do not lay eggs typically have teeth; scorpions do not lay eggs, yet lack teeth: another reason for considering them anomalous.
The platypus and seasnake are considered anomalous because animals typically give milk or lay eggs (i.e. one or the other, not both); a platypus does both, while a seasnake does neither. 

\admercs{} also identifies an anomalous context of flying hairy animals with members: honeybee, housefly, wasp, moth, vampire bat and fruitbat.
The combination of ``hair'' and ``airborne'' attributes both being true is apparently rare in the dataset, and covers a highly diverse range of animals, implying that each animal on its own is somewhat atypical among its peers for having this combination. 
This anomalous context is the highest-scoring in this dataset, but its anomaly score is still relatively low. 


\begin{figure}[t]
        \centering
        \begin{tabular}{l@{~~~}l@{~~~}l}
                    \toprule
                   
                    \textit{Animals that ... } & \textit{typically ...} & \textit{but a scorpion ...} \\ 
                    \midrule
                    do not lay eggs & have teeth & does not have teeth. \\
                    do not lay eggs and are not aquatic & give milk & does not give milk.\\
        	        do not give milk & lay eggs & does not lay eggs. \\
        	        
        	       have no backbone& have no tail & has a tail.\\
        	    
        	      \bottomrule
        \end{tabular}
    
    
    \caption{Four explanations generated by \admercs{} to explain why a scorpion is anomalous in the Zoo dataset.}
    \label{fig:zoo_illustration}
\end{figure}

\section{Experimental Evaluation}\label{s:experiments}
Our experiments 
answer the following research questions:
\begin{itemize}
    \item [\textbf{Q1}] Is \admercs{} competitive with the state-of-the-art?
    \item [\textbf{Q2}] Is \admercs{} capable of:
    \begin{itemize}
        \item [\textbf{Q2.1}] finding relevant subspaces?
        \item [\textbf{Q2.2}] finding relevant contexts?
        \item [\textbf{Q2.3}] detecting accidental inliers?  
    \end{itemize}
\end{itemize}
We ask \textbf{Q1} to ensure that, generally speaking, \admercs{} is a competitive all-round anomaly detector.
Then, \textbf{Q2} systematically verifies specific, desirable properties that we tried to embed in \admercs{}.
All experiments follow the same structure.
We state a \textit{hypothesis}, describe \textit{datasets} and discuss the \textit{limitations} of that particular setup. 
Then, we report and discuss \textit{results} and draw \textit{conclusions}. 

Each of the following subsections is devoted to one of the experiments. 
First, in subsection \ref{ss:general}, we test general anomaly detection performance. 
Second, in subsection \ref{ss:subspace} and \ref{ss:context}, we test the subspace and context aspect of \admercs{} in isolation followed by subsection \ref{ss:subspace+context} where we consider subspace and context simultaneously. 
Finally, in subsection \ref{ss:anomalous_context}, we test \admercs{}' capability to detect accidental inliers. 
Common to all these experiments are the \textit{algorithms}, the \textit{evaluation metrics} and the \textit{hyperparameter tuning}.
Additionally, in the supplementary material, we provide characteristics of all datasets, the parameter grids and selected parameters of our gridsearch, the dataset-by-dataset performance of each algorithm, and parameter sensitivity plots for \admercs{}.

\paragraph{Algorithms.}
We compare \admercs{} to a representative sample of state-of-the-art anomaly detectors\footnote{We use \lof{}, \knn{} and \hbos{} from \texttt{pyod}\cite{zhao2019}; \iforest{} from \texttt{scikit-learn}\cite{scikit-learn} and \hics{} from \texttt{ELKI} (\texttt{github.com/elki-project/elki}).
The \also{} implementation is our own, using \texttt{scikit-learn} decision trees.}:
\lof{}\cite{Breunig2000SIGMOD}  and \knn{}\cite{dudani1976distance}, both nearest-neighbor approaches;
\hbos{}\cite{goldstein2012histogram}, \iforest{}\cite{Liu2008ICDM} and \hics{}\cite{Keller2012ICDE} as subspace methods;
and \also{}\cite{Paulheim2015MLJ} with regression trees as residual-based approach. 
Furthermore, \lof{}, \knn{} and \iforest{} emerge as top performers in recent and extensive empirical evaluations\cite{Campos2016DMKD,Emmott2015}.

\paragraph{Evaluation Metrics.}
We measure performance by two common metrics in AD\cite{Aggarwal2017OA}: the \textit{area under the receiver operating characteristic curve} ($\mathit{AUC}$) and the \textit{average precision} ($AP$).
Per experiment, we report an algorithm's average performance and rank, along with a \textit{critical distance} \criticaldist{} determined by a Nemenyi post-hoc test\cite{demsar2006statistical} with significance level $p=0.05$.
A lower rank means algorithm $A$ outperforms algorithm $B$. 
A difference between two ranks is significant it it exceeds the critical distance \criticaldist{}.
If there are multiple versions (due to subsampling) of the same dataset, performance is averaged across versions prior to calculation of the ranks.

\paragraph{Hyperparameter Tuning.} 
Per algorithm and experiment, we determine one set of hyperparameters using a grid search.
We select the hyperparameters that yield the highest average $AUC$  across a representative sample\footnote{For Campos and CamposHD, the first subsample version of each dataset\cite{Campos2016DMKD}; for the HiCS benchmark and Synth-C\&S, the first dataset of each dimensionality; and for Synth-C and Synth-I, we subsample 10 out of the 30 datasets.} of the datasets of an experiment (to keep the execution time under control).

\begin{figure}[t]
    \small
    \centering
    \includegraphics[width=0.8\columnwidth]{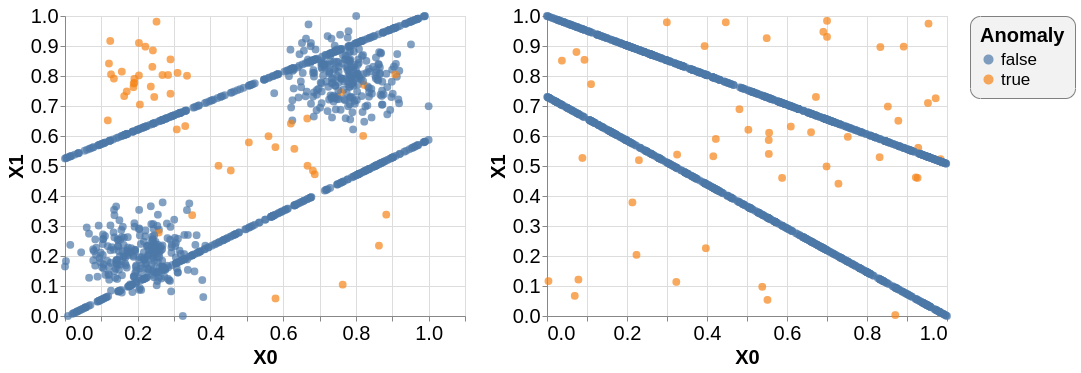}
	\caption{
	Examples of simple 2D patterns with obvious anomalies, as used in construction of our synthetic benchmarks: Synth-C, Synth-C\&S and Synth-I. 
	The anomalies are local outliers: breaking a typical pattern in the dataset, but without any individual attribute having a value that is atypical with respect to the full dataset.
	}
	\label{f:lcr-benchmark}
\end{figure}

\begin{table}[t]
\footnotesize
\caption{Experimental results.
Best average rank and best average performance are underlined. 
Ranks in bold do not differ significantly from the best rank.
For each experiment, \admercs{} is the top performer or within \criticaldist{} of the top performer.}
\label{t:sota-results}
\centering
\addtolength{\tabcolsep}{1pt}
\resizebox{\linewidth}{!}{
\begin{tabular}{@{}l|cccc|cccc|cccc|cccc|cccc| cccc@{}}
\toprule
{} & \multicolumn{4}{c|}{Campos} & \multicolumn{4}{c|}{CamposHD} & \multicolumn{4}{c|}{HiCS} & \multicolumn{4}{c|}{Synth-C} & \multicolumn{4}{c|}{Synth-C\&S} & \multicolumn{4}{c}{Synth-I} \\
{} & \multicolumn{4}{c|}{\textit{Literature}} & \multicolumn{4}{c|}{\textit{Irrelevant Dimensions}} & \multicolumn{4}{c|}{\textit{Multiple Subspaces}} & \multicolumn{4}{c|}{\textit{Context}} & \multicolumn{4}{c|}{\textit{Context+Subspace}}& \multicolumn{4}{c}{\textit{Accidental Inliers}} \\
{} & \multicolumn{2}{c}{AUC} & \multicolumn{2}{c|}{AP} & \multicolumn{2}{c}{AUC} & \multicolumn{2}{c|}{AP} & \multicolumn{2}{c}{AUC} & \multicolumn{2}{c|}{AP} & \multicolumn{2}{c}{AUC} & \multicolumn{2}{c|}{AP} & \multicolumn{2}{c}{AUC} & \multicolumn{2}{c|}{AP} & \multicolumn{2}{c}{AUC} & \multicolumn{2}{c}{AP} \\
{} &               avg &            rank &        avg &            rank &         avg &            rank &        avg &            rank &        avg &            rank &        avg &            rank &        avg &            rank &        avg &            rank &        avg &            rank &        avg &           rank &        avg &            rank &       avg &            rank \\
\midrule
\admercs{} &              0.78 &       \tb{3.98} &       0.31 &        \tb{3.9} &   \tu{0.77} &       \tb{3.58} &       0.27 &       \tb{3.64} &  \tu{0.99} &  \tb{\tu{1.1}} &  \tu{0.89} &  \tb{\tu{1.24}} &  \tu{0.96} &        \tb{2.4} &  \tu{0.83} &       \tb{2.37} &  \tu{0.96} &  \tb{\tu{1.37}} &  \tu{0.82} &  \tb{\tu{1.23}} &  \tu{0.93} &  \tb{\tu{1.5}} &  \tu{0.79} &  \tb{\tu{1.4}} \\
\also{}    &              0.68 &            5.51 &        0.2 &            5.28 &         0.7 &       \tb{3.98} &       0.24 &       \tb{4.14} &       0.96 &       \tb{2.1} &       0.77 &       \tb{2.14} &       0.88 &            4.53 &       0.61 &            4.47 &       0.95 &       \tb{1.67} &       0.73 &       \tb{1.87} &       0.88 &           3.53 &       0.54 &           3.37 \\
\hbos{}    &              0.78 &       \tb{3.99} &       0.31 &       \tb{3.91} &   \tu{0.77} &  \tb{\tu{3.08}} &  \tu{0.29} &        \tb{3.4} &       0.81 &           3.67 &       0.29 &            4.57 &       0.47 &            6.93 &       0.05 &            6.93 &        0.5 &            5.93 &       0.05 &            6.03 &       0.85 &           4.67 &       0.41 &           4.83 \\
\hics{}    &              0.76 &       \tb{4.36} &       0.27 &       \tb{4.16} &        0.71 &       \tb{3.38} &       0.22 &  \tb{\tu{3.39}} &        0.8 &            4.0 &        0.6 &       \tb{2.76} &       0.88 &            4.05 &       0.68 &            3.73 &       0.72 &       \tb{2.97} &       0.39 &             2.9 &       0.78 &           5.15 &       0.38 &           4.82 \\
\iforest{} &              0.79 &       \tb{3.36} &  \tu{0.33} &       \tb{3.65} &        0.73 &        \tb{3.9} &  \tu{0.29} &       \tb{3.73} &       0.71 &           6.38 &        0.2 &            6.67 &       0.93 &            4.23 &       0.57 &             5.2 &       0.55 &            5.73 &       0.07 &            5.53 &        0.9 &      \tb{3.03} &       0.53 &           3.67 \\
\knn{}     &          \tu{0.8} &  \tb{\tu{3.04}} &  \tu{0.33} &  \tb{\tu{3.06}} &        0.69 &            5.09 &       0.25 &       \tb{4.88} &       0.74 &           4.86 &       0.26 &            5.14 &  \tu{0.96} &  \tb{\tu{1.87}} &        0.8 &  \tb{\tu{1.93}} &       0.57 &            5.07 &        0.1 &             5.4 &       0.76 &           4.37 &       0.34 &            4.8 \\
\lof{}     &              0.79 &       \tb{3.76} &       0.27 &       \tb{4.04} &        0.68 &       \tb{4.98} &       0.24 &       \tb{4.82} &       0.72 &            5.9 &       0.24 &            5.48 &       0.88 &            3.98 &       0.68 &       \tb{3.37} &       0.57 &            5.27 &        0.1 &            5.03 &       0.71 &           5.75 &       0.32 &           5.12 \\
\midrule
\criticaldist{}  & \multicolumn{4}{c|}{1.88} &\multicolumn{4}{c|}{1.96} &  \multicolumn{4}{c|}{1.96} &  \multicolumn{4}{c|}{1.64} & \multicolumn{4}{c|}{1.64} & \multicolumn{4}{c}{1.64}   \\

\bottomrule
\end{tabular}}
\end{table}

\subsection{General Performance}\label{ss:general}
To test whether \admercs{} is competitive with state-of-the-art AD-algorithms (\textbf{Q1}), we use $23$ real-world datasets from the AD literature: the \textit{Campos benchmark} (Campos et al.\cite{Campos2016DMKD})\footnote{
For each dataset we use the normalized versions without duplicates with a $5\%$ contamination. 
For datasets where a version with $5\%$ contamination is absent, we perform our own subsampling.}. 
The main limitation of this experiment is that the characteristics of the anomalies in these datasets are unknown and beyond our control\cite{Emmott2015,Steinbuss2020}. 
Therefore, this experiment does not allow us to draw any definitive conclusions with regards to an algorithm's capability to detect anomalies in subspaces, contexts, etc... 

\paragraph{Results.}
The leftmost column of Table~\ref{t:sota-results} summarizes our results.
Top performers on the Campos benchmark are \knn{}, \iforest{} and \lof{}, closely followed by \hbos{}, \admercs{} and \hics{}. 
Note that, among these methods, \textit{no significant performance difference} exists.
\also{}, however, does record a significantly lower performance.
Additionally, \hbos{}' impressive performance on this experiment indicates that most of the anomalies in the Campos benchmark are, in fact, global outliers.

\paragraph{Conclusion.}
Generally speaking, \admercs{} is a competitive anomaly detector, since it performs at par with the state-of-the art.
\hbos{}' surprising performance is something to keep in mind when interpreting the results of this experiment; global outliers are comparatively easy to detect, and consequently, this benchmark offers a quite limited perspective on the true capabilities of an anomaly detector.

\subsection{Anomaly Detection with Subspaces}\label{ss:subspace}
In this experiment, we aim to show that \admercs{} can find the right subspace(s) to do anomaly detection (\textbf{Q2.1}).
First, to test robustness to uninformative dimensions, we built CamposHD, a version of the Campos benchmark with synthetic irrelevant dimensions: for a dataset with $n$ attributes, we add $4n$ uninformative attributes uniformly sampled from $[0,1]$.\footnote{
Except datasets InternetAds and Arrhythmia, which are already high-dimensional.}
Second, the HiCS benchmark (Keller et al.\cite{Keller2012ICDE}) tests an algorithm's capability to find useful subspaces for AD: it consists of seven synthetic high-dimensional datasets with dense clusters in low-dimensional subspaces and anomalies that fall outside of these clusters.
Both collections of datasets are useful to investigate the subspace aspect of the AD-problem in particular. 

\paragraph{Results.}
The second and third column of Table~\ref{t:sota-results} summarize our results.
\lof{} and \knn{} struggle when uninformative dimensions and subspaces come into play: their performance drops between Campos and CamposHD, and they are amongst the worst performers on HiCS.
\iforest{}' random subspaces are somewhat robust to uninformative dimensions, but are inadequate to detect anomalies scattered across multiple subspaces.
\hbos{} (which only detects global anomalies) becomes the top performer on CamposHD, its performance is almost unchanged between Campos and CamposHD meaning that the algorithm is robust to the uninformative dimensions. 
However, \hbos{} struggles to detect HiCS' anomalies hidden in non-trivial subspaces.
\hics{}, being an algorithm designed for subspace-AD, records strong performances on both experiments.
\also{} becomes a much more attractive option with subspaces coming into play.
Finally, \admercs{} claims the top spot on the HiCS benchmark and lies within \criticaldist{} of the top performers (\hbos{} in terms of AUC and \hics{} in terms of AP) on CamposHD.

\paragraph{Conclusion.}
\admercs{} performs at-par with the top algorithms on CamposHD and is the top performer on the HiCS benchmark, which indicates that it handles subspaces effectively.

\subsection{Anomaly Detection with Contexts}\label{ss:context}
To test whether \admercs{} can detect local outliers by identifying the right context(s) to do anomaly detection (\textbf{Q2.2}), we construct $30$ simple $2D$ patterns with obvious anomalies (two example patterns are shown in Fig.~\ref{f:lcr-benchmark}). 
The patterns and anomalies are chosen such that anomalies are invisible in the marginal distribution of any single attribute, which means that choosing proper contexts is necessary for their detection. 
We call this synthetic benchmark Synth-C. 
Proficiency on this benchmark indicates an algorithm's capability to identify appropriate contexts for anomaly detection.

\paragraph{Results.}
The fourth column of Table~\ref{t:sota-results} summarizes our results.
When it comes to finding proper contexts for these simple 2D datasets, \knn{} performs best but \admercs{} and \lof{} are within critical distance of \knn{}.  
\hbos{} consistently ranks last on this experiment, which is of course due to the fact that individual marginals carry no information here.
\also{} appears to underperform too; this happens because it cannot capture the non-functional dependencies that are present in some of the datasets.

\paragraph{Conclusion.}
\admercs{} is able to identify adequate contexts to spot the anomalies on a collection of 2D benchmark datasets.

\subsection{Anomaly Detection with Subspaces and Contexts}\label{ss:subspace+context}
To test whether \admercs{} simultaneously finds good subspace(s) and the context(s) (\textbf{Q2.1} \& \textbf{Q2.2}), we introduce Synth-C\&S, a collection of $30$ benchmark datasets where successful AD requires both context and subspaces. 
Each dataset here contains $5$ relevant $2D$ subspaces, each subspace containing a simple $2D$ pattern with obvious anomalies (exactly like those used in Synth-C), and a varying number of irrelevant dimensions randomly sampled from $[0,1]$.

\paragraph{Results.}
The fifth column of Table~\ref{t:sota-results} summarizes our results.
When the detection of anomalies requires both finding relevant subspaces as well as finding adequate contexts, \admercs{} is consistently the top-ranked method.
Within \criticaldist{}, we encounter \also{}.
\hics{}' performance is still reasonable and captures third place. 
All other algorithm struggle under the conditions imposed by this experiment, as illustrated by a large performance gap. 

\paragraph{Conclusion.}
If successful AD requires both finding relevant subspaces as well as use of proper contexts, \admercs{} is consistently the top-ranked algorithm.

\subsection{Anomaly Detection with Accidental Inliers}\label{ss:anomalous_context}
To actually verify that \admercs{} is able to detect accidental inliers (\textbf{Q2.3}), we introduce Synth-I, a collection of $30$ datasets where some anomalies can only be detected by realizing they belong to an anomalous context. 
Each dataset contains two subspaces: first, a simple $2D$ pattern with clear local outliers (again, exactly like those used in Synth-C) and second, a $2D$ subspace with a few clusters including at least one \textit{anomalous cluster}: a cluster where the majority of instances are local outliers in the first $2D$ subspace. 
However, a minority of instances in the anomalous cluster will be perfectly normal in the first subspace; they are anomalous because, in the second subspace, they resemble other anomalies.

\paragraph{Results.}
The last column of Table~\ref{t:sota-results} summarizes our results.
If one wants to detect instances similar to known anomalies, \admercs{}' detection of anomalous contexts works: \admercs{} is the best performer here.

\paragraph{Conclusion.}
\admercs{} is able to detect accidental inliers. 

\subsection{Summary} 
We conclude that \admercs{} is a competitive anomaly detector in general, regardless of what kind of anomalies are present (\textbf{Q1}).
More importantly, \admercs{} finds and exploits relevant subspaces (\textbf{Q2.1}) and contexts (\textbf{Q2.2}) and is especially effective when both are needed simultaneously.
Finally,  \admercs{} can detect accidental inliers (\textbf{Q2.3}).

\section{Conclusions}
In this paper, we presented \admercs{}, a novel anomaly detection approach that is unique in that it exploits both normal and anomalous patterns between attributes to detect and explain anomalies. 
Due to this property, it can detect accidental inliers, a type of anomalies not supported by the current state-of-the-art, and it can provide both positive and negative explanations for why something is an anomaly. 
We have shown experimentally that \admercs{} is competitive with the state of the art on known anomaly detection benchmarks and a top performer on high-dimensional datasets because \admercs{} effectively exploits subspaces and contexts.
Finally, we identified a too strong focus on global outliers in known benchmark datasets and contributed a novel set of intuitive and interpretable benchmarks with local outliers.



\section*{Acknowledgments}
This research received funding from the Flemish Government (AI Research Program) and the European Research Council (ERC) under the European Union’s Horizon 2020 research and innovation programme (grant agreement No. 694980 ``\texttt{SYNTH}: Synthesising Inductive Data Models'').

\bibliographystyle{splncs04}  
\bibliography{references}

\end{document}


\maketitle              

\appendix
\section{Gridsearch Details}\label{section:gridsearch}
As mentioned in the text, in order to keep the execution time of our experiments under control, the gridsearch was done on a representative subsample of all datasets: for Campos and CamposHD, one subsample variation of each dataset; for HiCS and Synth-C\&S, one dataset of each dimensionality; for Synth-C and Synth-I, a subsample of 10 out of 30 datasets. 
In our experience, and to the best of our knowledge, this subsample of datasets is indeed representative; the differences between the parameters ultimately selected by our gridsearch vs. those selected by a gridsearch on truly all datasets can be expected to be negligible for all practical purposes, whereas runtime can be expected to increase by an order of magnitude.

\noindent
The parameter grids used for the gridsearch are shown below and the selected parameters are shown in Table \ref{table:best_parameters}.
\begin{itemize} 
    \item \admercs{} has parameters: the tree learning parameters, $\rho$, $\gamma_\delta$ and $\gamma_\lambda$. 
    To learn regression trees we use the friedman MSE criterion, to learn classification tree we use gini impurity.
    \begin{itemize}
        \item max depth $= 10$ (fixed) i.e. max tree depth.
        \item min samples leaf $= [0.005, 0.02, 0.05, 0.1]$ i.e. the minimal number of instances required in each leaf as a percentage of the number of instances in the dataset.
        \item min impurity decrease $= [ 0.001, 0.05, 0.2, 0.5]$ i.e. the minimal impurity decrease a split needs to have before it is considered a valid candidate.
        \item $\rho = [0.7, 0.9]$ 
        \item $\gamma_\lambda =  [ 1.0, 0.9, 0.5]$ 
        \item $\gamma_\delta = [ 1.0, 0.9, 0.7, 0.2]$ 
    \end{itemize}
    \item \also{} only needs tree learning parameters, we used the same grid as for \admercs{} but with several max depth options. 
    We only used regression trees with friedman MSE as a splitting criterion.
    \begin{itemize}
        \item max depth $= [4,10,16]$ 
        \item min samples leaf $= [0.005, 0.02, 0.05, 0.1]$ 
        \item min impurity decrease $= [ 0.001, 0.05, 0.2, 0.5]$ 
    \end{itemize}
    \item \hbos{} only needs the number of bins to use for its histograms. 
    \begin{itemize}
        \item number of bins $= [5,10,15,20,25,30,35,40]$
    \end{itemize}
    \item \hics{} only needs the the number of neighbors $k$ that is used to run \lof{} in each subspace. 
    \hics{} is run with less values for $k$ than \knn{} and \lof{} because \hics{}' execution time is considerably larger. 
    \begin{itemize}
        \item $k = [5, 7, 10, 15, 20, 25, 30, 35, 40, 45, 50]$
    \end{itemize}
    \item \iforest{} needs the number of trees in the ensemble and the maximum number of samples it uses to learn each tree (relative to the number of instances in the data).
    \begin{itemize}
        \item n estimators $= [100,200,300,400,500]$
        \item max samples $ = [0.1, 0.25, 0.5, 0.75,1.0]$
    \end{itemize}
    \item \knn{} only needs the number of neighbors $k$.
    \begin{itemize}
        \item $k = [3:50]$ all values between 3 and 50 were used .
    \end{itemize}
    \item \lof{} only needs the number of neighbors $k$.
    \begin{itemize}
        \item $k = [3:50]$ all values between 3 and 50 were used.
    \end{itemize}
        
\end{itemize}

\begin{table}[h]
    \caption{The optimal parameters resulting from our gridsearch. For each benchmark, we selected the set with the highest average $AUC$ over all datasets in the benchmark.
    If we do not mention a parameter here and a fixed value is not mentioned in the grid configuration (cf. appendix \ref{section:gridsearch}), we rely on the default value of the algorithm's own implementation.}
    \label{table:best_parameters}
    \resizebox{\linewidth}{!}{
    \begin{tabular}{lll}
\toprule
Algorithm & Benchmark &            Best parameters                                                                                                               \\
\midrule
\admercs{} & Campos &    min\_samples\_leaf = 0.05, min\_impurity\_decrease = 0.05, $\rho$ = 0.9, $\gamma_\lambda$ = 0.5, $\gamma_\delta$ = 0.7 \\
       & CamposHD &      min\_samples\_leaf = 0.1, min\_impurity\_decrease = 0.2, $\rho$ = 0.7, $\gamma_\lambda$ = 1.0, $\gamma_\delta$ = 0.7 \\
       & HiCS &     min\_samples\_leaf = 0.02, min\_impurity\_decrease = 0.5, $\rho$ = 0.9, $\gamma_\lambda$ = 1.0, $\gamma_\delta$ = 1.0 \\
       & Synth-C &   min\_samples\_leaf = 0.02, min\_impurity\_decrease = 0.001, $\rho$ = 0.7, $\gamma_\lambda$ = 0.5, $\gamma_\delta$ = 0.2 \\
       & Synth-C\&S &   min\_samples\_leaf = 0.05, min\_impurity\_decrease = 0.001, $\rho$ = 0.9, $\gamma_\lambda$ = 1.0, $\gamma_\delta$ = 1.0 \\
       & Synth-I &  min\_samples\_leaf = 0.005, min\_impurity\_decrease = 0.001, $\rho$ = 0.9, $\gamma_\lambda$ = 0.5, $\gamma_\delta$ = 0.9 \\
\also{} & Campos &                                             max\_depth = 10.0, min\_samples\_leaf = 0.02, min\_impurity\_decrease = 0.001 \\
       & CamposHD &                                              max\_depth = 10.0, min\_samples\_leaf = 0.05, min\_impurity\_decrease = 0.05 \\
       & HiCS &                                              max\_depth = 16.0, min\_samples\_leaf = 0.1, min\_impurity\_decrease = 0.001 \\
       & Synth-C &                                               max\_depth = 16.0, min\_samples\_leaf = 0.02, min\_impurity\_decrease = 0.2 \\
       & Synth-C\&S &                                               max\_depth = 10.0, min\_samples\_leaf = 0.02, min\_impurity\_decrease = 0.2 \\
       & Synth-I &                                               max\_depth = 16.0, min\_samples\_leaf = 0.02, min\_impurity\_decrease = 0.2 \\
\hbos{} & Campos &                                                                                                              n\_bins = 10 \\
       & CamposHD &                                                                                                              n\_bins = 10 \\
       & HiCS &                                                                                                              n\_bins = 20 \\
       & Synth-C &                                                                                                              n\_bins = 10 \\
       & Synth-C\&S &                                                                                                              n\_bins = 30 \\
       & Synth-I &                                                                                                               n\_bins = 5 \\
\hics{} & Campos &                                                                                                                    k = 45 \\
       & CamposHD &                                                                                                                    k = 30 \\
       & HiCS &                                                                                                                    k = 20 \\
       & Synth-C &                                                                                                                    k = 25 \\
       & Synth-C\&S &                                                                                                                    k = 15 \\
       & Synth-I &                                                                                                                    k = 50 \\
\iforest{} & Campos &                                                                                   n\_estimators = 100, max\_samples = 0.5 \\
       & CamposHD &                                                                                   n\_estimators = 300, max\_samples = 1.0 \\
       & HiCS &                                                                                   n\_estimators = 500, max\_samples = 1.0 \\
       & Synth-C &                                                                                   n\_estimators = 500, max\_samples = 1.0 \\
       & Synth-C\&S &                                                                                   n\_estimators = 100, max\_samples = 1.0 \\
       & Synth-I &                                                                                  n\_estimators = 100, max\_samples = 0.75 \\
\knn{} & Campos &                                                                                                                     k = 9 \\
       & CamposHD &                                                                                                                    k = 10 \\
       & HiCS &                                                                                                                     k = 4 \\
       & Synth-C &                                                                                                                     k = 4 \\
       & Synth-C\&S &                                                                                                                     k = 3 \\
       & Synth-I &                                                                                                                     k = 7 \\
\lof{} & Campos &                                                                                                                    k = 29 \\
       & CamposHD &                                                                                                                    k = 29 \\
       & HiCS &                                                                                                                    k = 22 \\
       & Synth-C &                                                                                                                    k = 24 \\
       & Synth-C\&S &                                                                                                                     k = 4 \\
       & Synth-I &                                                                                                                    k = 50 \\
\bottomrule
\end{tabular}}
\end{table}

\section{Dataset Characteristics}
Dataset statistics for the Campos, CamposHD and HiCS benchmarks are shown in Table \ref{table:campos_statistics}, \ref{table:campusHD_statistics} and \ref{table:hics_statistics} respectively. 
The Campos and CamposHD have 10 subsampling variations of each dataset, except for Wilt which has only one version. 
The statistics for our own synthetic benchmarks are as follows: 
The Synth-C benchmark contains 30 datasets with 1000 instances, a dimensionality of 2 and a contamination of 0.05. 
The Synth-C\&S benchmark contains 3 datasets of each dimensionality in $[10,20,30,40,50,60,70,80,90,100]$ (30 in total), each of these has 1000 instances and a contamination of 0.05. 
Finally the Synth-I benchmark contains 30 datasets with 1000 instances, a dimensionality of 4 and a contamination of 0.05.

\begin{table}[h!]
    \centering
    \caption{Dataset statistics on the datasets we use from the Campos benchmark (221 datasets). All datasets have 10 different subsampling versions except for Wilt which only has one.}
    \label{table:campos_statistics}
   \begin{tabular}{lrrc}
        \toprule
        Dataset &  \#Instances &  \#Dimensions & Contamination \\
        
        \midrule
        ALOI             &       30160 &           27 &          0.05 \\
        Annthyroid       &        6942 &           21 &          0.05 \\
        Arrhythmia       &         256 &          259 &          0.05 \\
        Cardiotocography &        1734 &           21 &          0.05 \\
        Glass            &         180 &            7 &          0.05 \\
        HeartDisease     &         157 &           13 &          0.04 \\
        Hepatitis        &          70 &           19 &          0.04 \\
        InternetAds      &        1682 &         1555 &          0.05 \\
        Ionosphere       &         237 &           32 &          0.05 \\
        KDDCup99         &        4000 &           40 &          0.05 \\
        Lymphography     &         120 &           18 &          0.05 \\
        PageBlocks       &        5139 &           10 &          0.05 \\
        Parkinson        &          50 &           22 &          0.04 \\
        PenDigits        &         400 &           16 &          0.05 \\
        Pima             &         526 &            8 &          0.05 \\
        Shuttle          &         260 &            9 &          0.05 \\
        SpamBase         &        2661 &           57 &          0.05 \\
        Stamps           &         325 &            9 &          0.05 \\
        WBC              &         200 &            9 &          0.05 \\
        WDBC             &         200 &           30 &          0.05 \\
        WPBC             &         159 &           33 &          0.05 \\
        Waveform         &        2000 &           21 &          0.05 \\
        Wilt             &        4819 &            5 &          0.05 \\
        \bottomrule
    \end{tabular}
\end{table}

\begin{table}[h!]
    \centering
    \caption{Dataset statistics on the datasets in the CamposHD benchmark (201 datasets). All datasets have 10 different subsampling versions except for Wilt which only has one.}
    \label{table:campusHD_statistics}
   \begin{tabular}{lrrc}
        \toprule
        Dataset &  \#Instances &  \#Dimensions & Contamination \\
        
       \midrule
        ALOI             &       30160 &          135 &          0.05 \\
        Annthyroid       &        6942 &          105 &          0.05 \\
        Cardiotocography &        1734 &          105 &          0.05 \\
        Glass            &         180 &           35 &          0.05 \\
        HeartDisease     &         157 &           65 &          0.04 \\
        Hepatitis        &          70 &           95 &          0.04 \\
        Ionosphere       &         237 &          160 &          0.05 \\
        KDDCup99         &        4000 &          200 &          0.05 \\
        Lymphography     &         120 &           90 &          0.05 \\
        PageBlocks       &        5139 &           50 &          0.05 \\
        Parkinson        &          50 &          110 &          0.04 \\
        PenDigits        &         400 &           80 &          0.05 \\
        Pima             &         526 &           40 &          0.05 \\
        Shuttle          &         260 &           45 &          0.05 \\
        SpamBase         &        2661 &          285 &          0.05 \\
        Stamps           &         325 &           45 &          0.05 \\
        WBC              &         200 &           45 &          0.05 \\
        WDBC             &         200 &          150 &          0.05 \\
        WPBC             &         159 &          165 &          0.05 \\
        Waveform         &        2000 &          105 &          0.05 \\
        Wilt             &        4819 &           25 &          0.05 \\
\bottomrule
    \end{tabular}
\end{table}

\begin{table}[h!]
    \centering
    \caption{Dataset statistics on the datasets in the HiCS benchmark (21 datasets). }
    \label{table:hics_statistics}
   \begin{tabular}{lrrc}
        \toprule
        Dataset &  \#Instances &  \#Dimensions & Contamination \\
        
       \midrule
        synth\_multidim\_010\_000 &        1000 &           10 &          0.02 \\
        synth\_multidim\_010\_001 &        1000 &           10 &          0.02 \\
        synth\_multidim\_010\_002 &        1000 &           10 &          0.01 \\
        synth\_multidim\_020\_000 &        1000 &           20 &          0.03 \\
        synth\_multidim\_020\_001 &        1000 &           20 &          0.03 \\
        synth\_multidim\_020\_002 &        1000 &           20 &          0.04 \\
        synth\_multidim\_030\_000 &        1000 &           30 &          0.04 \\
        synth\_multidim\_030\_001 &        1000 &           30 &          0.04 \\
        synth\_multidim\_030\_002 &        1000 &           30 &          0.04 \\
        synth\_multidim\_040\_000 &        1000 &           40 &          0.05 \\
        synth\_multidim\_040\_001 &        1000 &           40 &          0.05 \\
        synth\_multidim\_040\_002 &        1000 &           40 &          0.06 \\
        synth\_multidim\_050\_000 &        1000 &           50 &          0.07 \\
        synth\_multidim\_050\_001 &        1000 &           50 &          0.07 \\
        synth\_multidim\_050\_002 &        1000 &           50 &          0.07 \\
        synth\_multidim\_075\_000 &        1000 &           75 &          0.11 \\
        synth\_multidim\_075\_001 &        1000 &           75 &          0.11 \\
        synth\_multidim\_075\_002 &        1000 &           75 &          0.11 \\
        synth\_multidim\_100\_000 &        1000 &          100 &          0.14 \\
        synth\_multidim\_100\_001 &        1000 &          100 &          0.12 \\
        synth\_multidim\_100\_002 &        1000 &          100 &          0.14 \\
        \bottomrule
    \end{tabular}
\end{table}

\section{Experimental Results per Dataset}
\begin{figure}[h!]
    \centering
    \includegraphics[width = 0.9\textwidth]{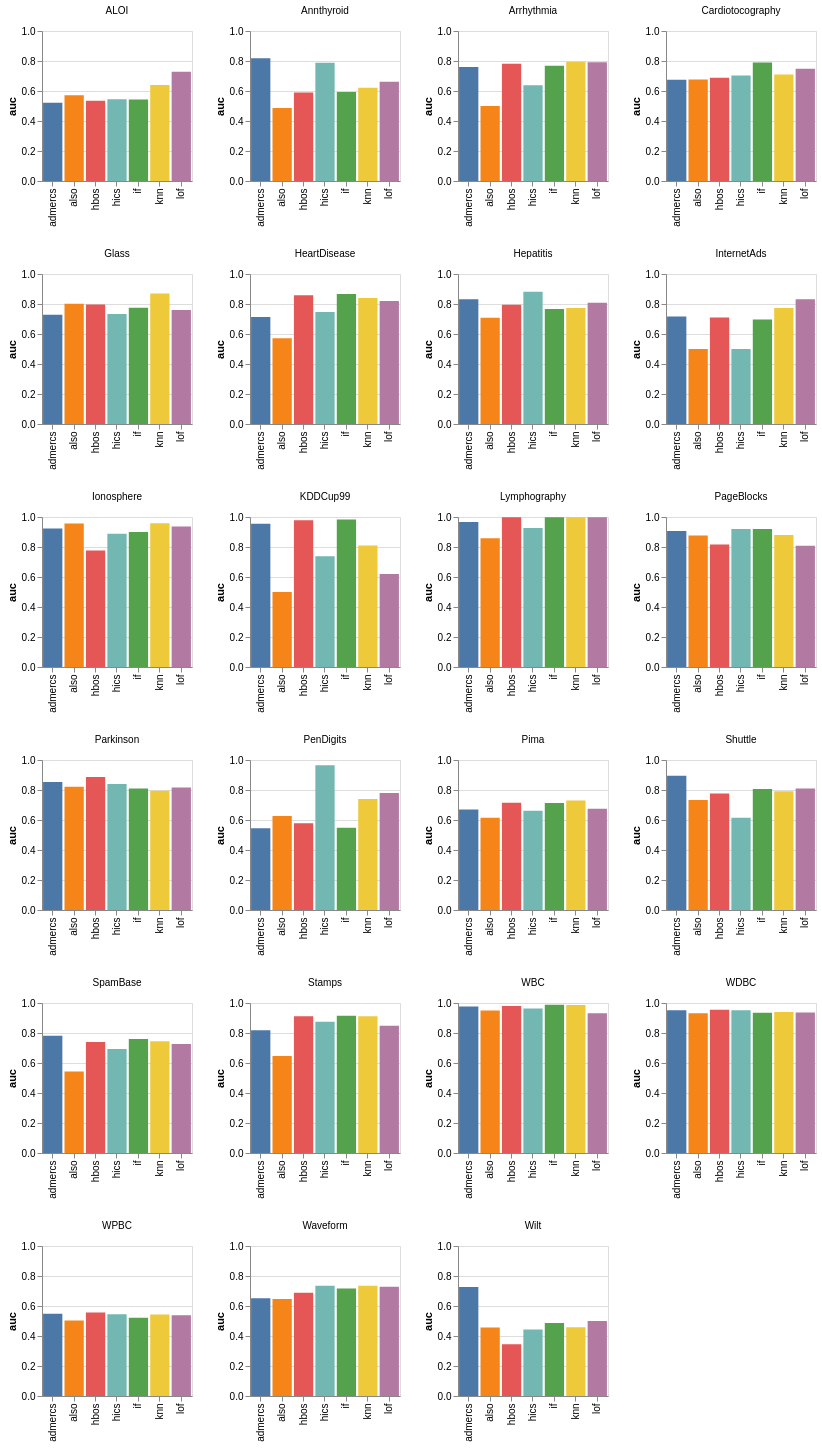}
    \caption{Individual AUC performance for each dataset and each algorithm (averaged over different subsample variations) for the Campos Benchmark with a single parameter set for all datasets.}
    \label{fig:auc_per_dataset_campos}
\end{figure}

\begin{figure}[h!]
    \centering
    \includegraphics[width = 0.9\textwidth]{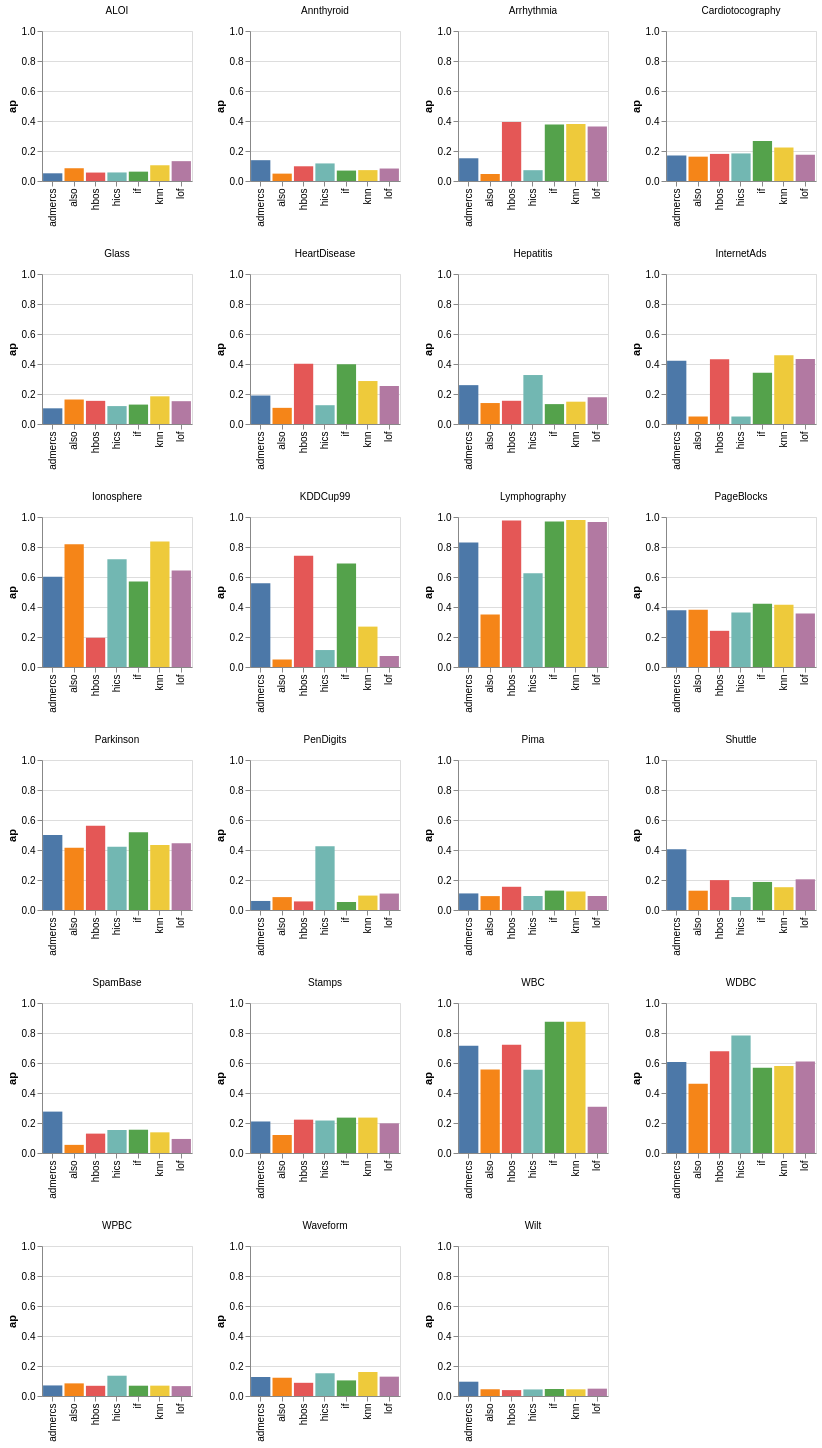}
    \caption{Individual AP performance for each dataset and each algorithm (averaged over different subsample variations) for the Campos benchmark with a single parameter set for all datasets.}
    \label{fig:ap_per_dataset_campos}
\end{figure}

\begin{figure}[h!]
    \centering
    \includegraphics[width = 0.9\textwidth]{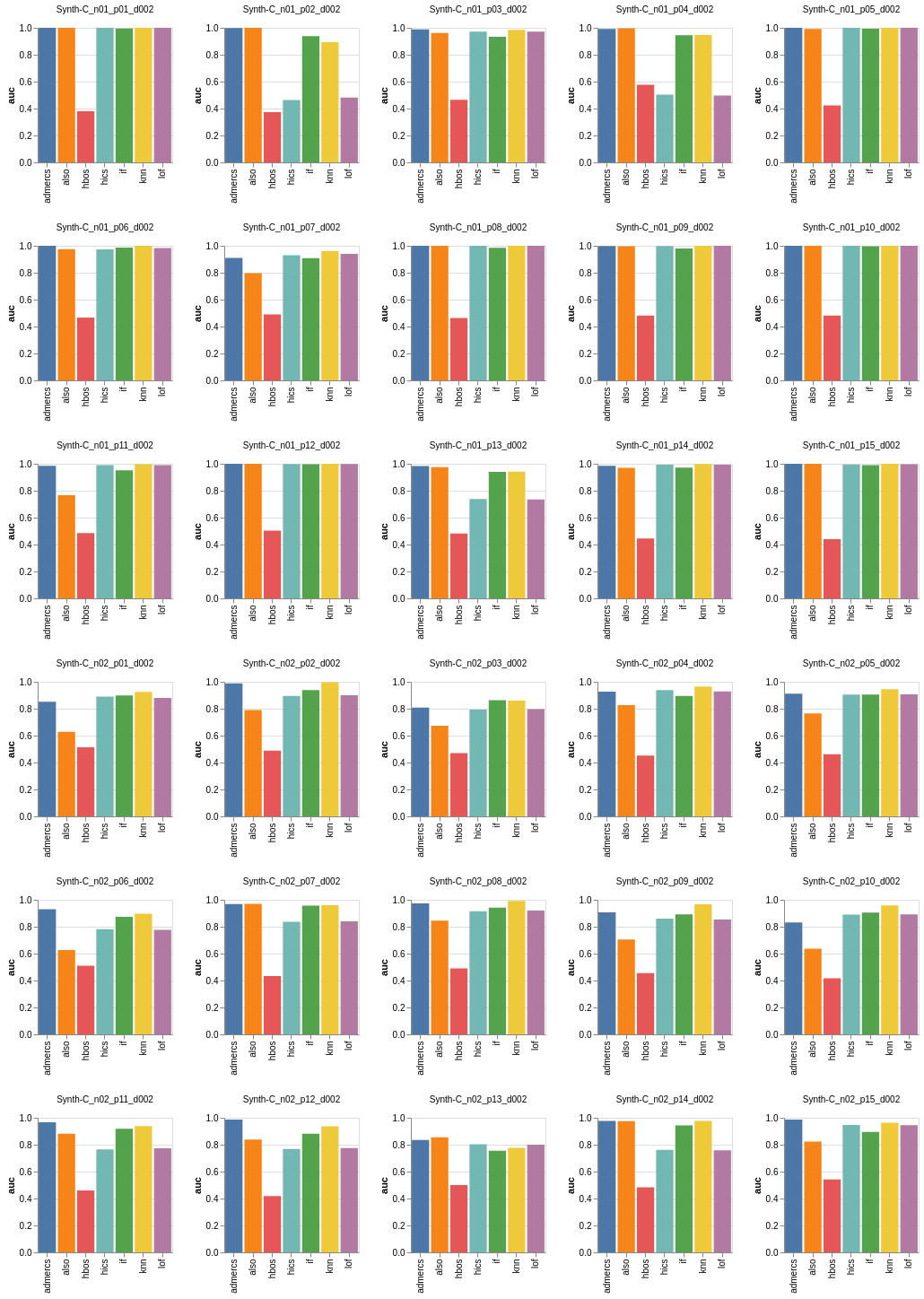}
    \caption{Individual AUC performance for each dataset and each algorithm for the Synth-C benchmark with a single parameter set for all datasets.}
    \label{fig:auc_per_dataset_context}
\end{figure}

\begin{figure}[h!]
    \centering
    \includegraphics[width = 0.9\textwidth]{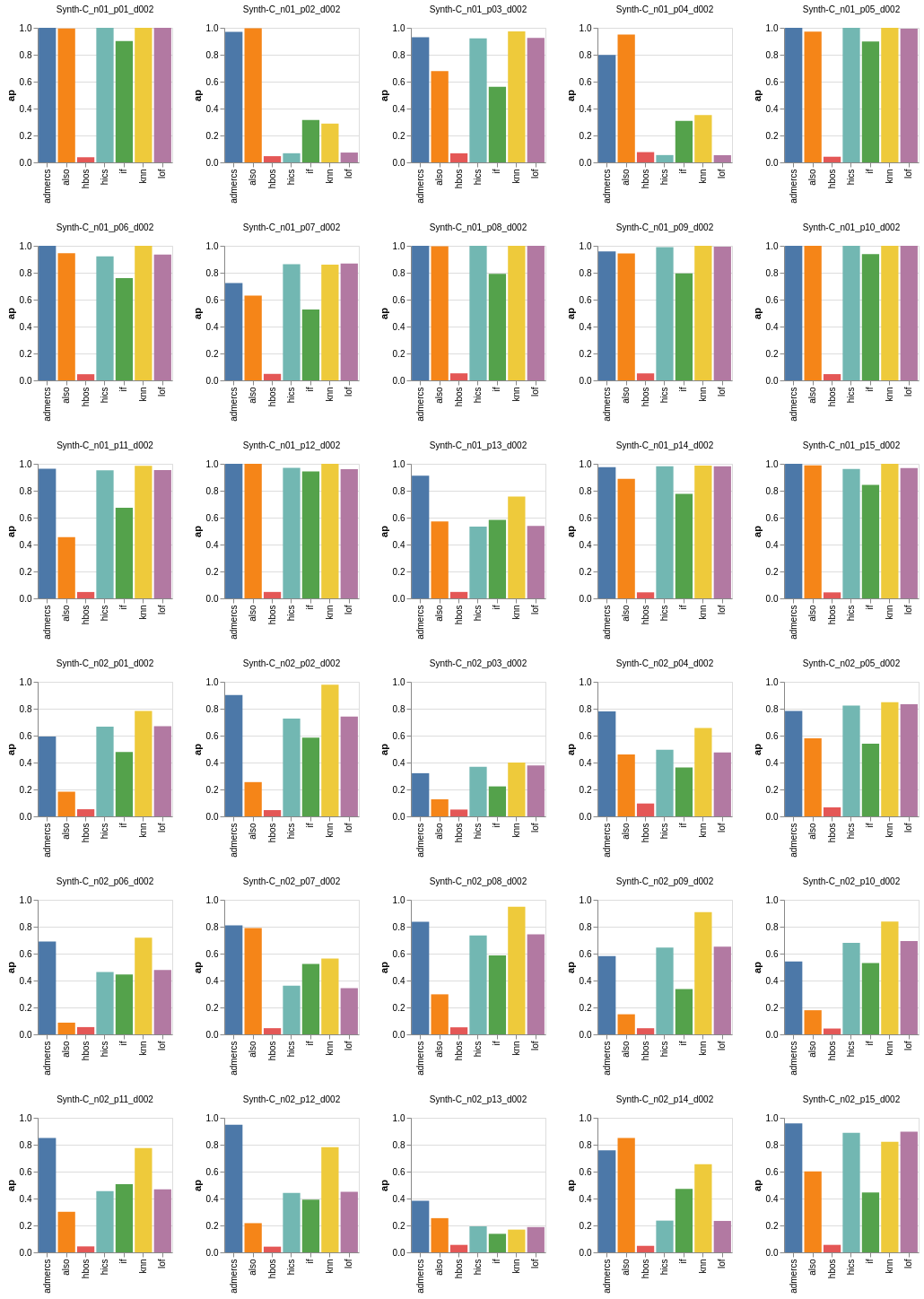}
    \caption{Individual AP performance for each dataset and each algorithm for the Synth-C benchmark with a single parameter set for all datasets.}
    \label{fig:ap_per_dataset_context}
\end{figure}

\begin{figure}[h!]
    \centering
    \includegraphics[width = 0.9\textwidth]{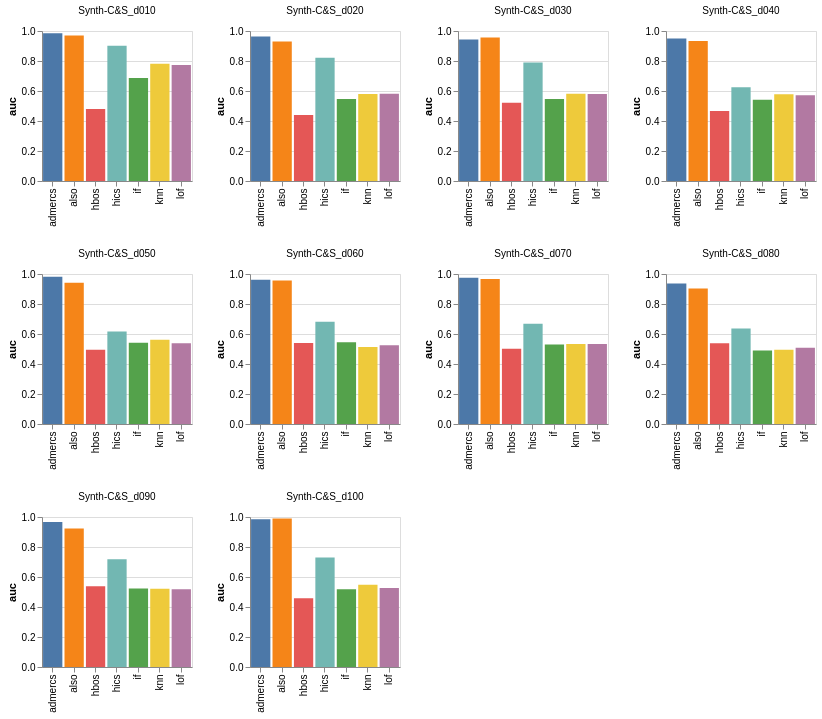}
    \caption{Average AUC performance for each dimensionality and each algorithm for the Synth-C\&S benchmark with a single parameter set for all datasets.}
    \label{fig:auc_per_dataset_contextsubspace}
\end{figure}

\begin{figure}[h!]
    \centering
    \includegraphics[width = 0.9\textwidth]{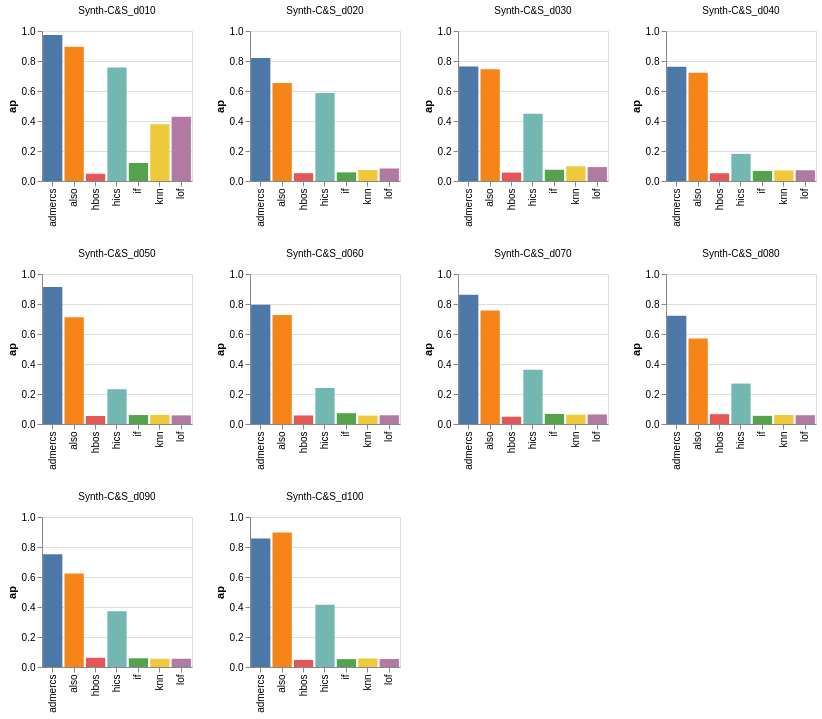}
    \caption{Average AP performance for each  dimensionality and each algorithm for the Synth-C\&S benchmark with a single parameter set for all datasets.}
    \label{fig:ap_per_dataset_contextsubspace}
\end{figure}

\begin{figure}[h!]
    \centering
    \includegraphics[width = 0.9\textwidth]{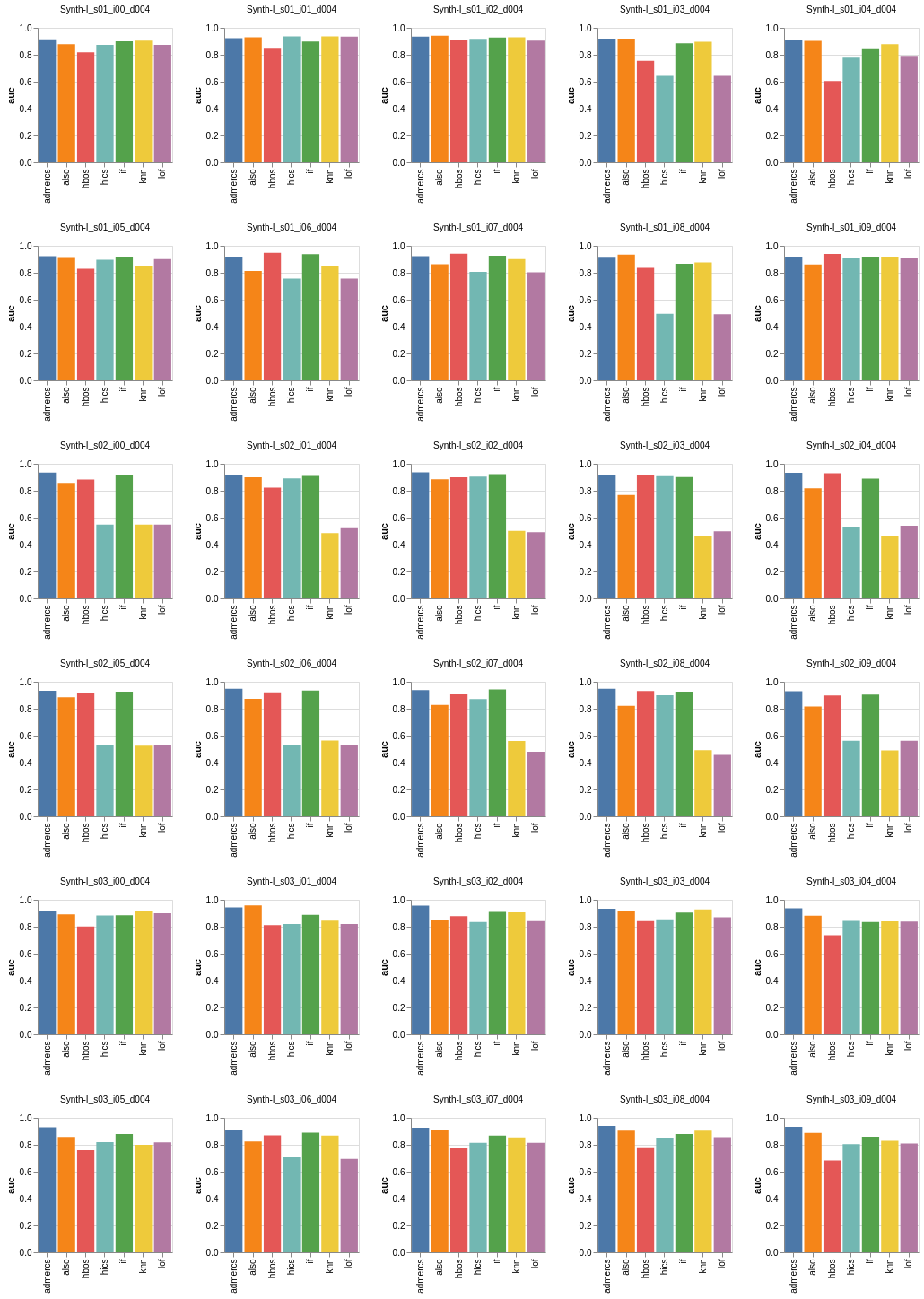}
    \caption{AUC performance for each dataset and each algorithm for the Synth-I benchmark with a single parameter set for all datasets.}
    \label{fig:auc_per_dataset_anomcontext}
\end{figure}

\begin{figure}[h!]
    \centering
    \includegraphics[width = 0.9\textwidth]{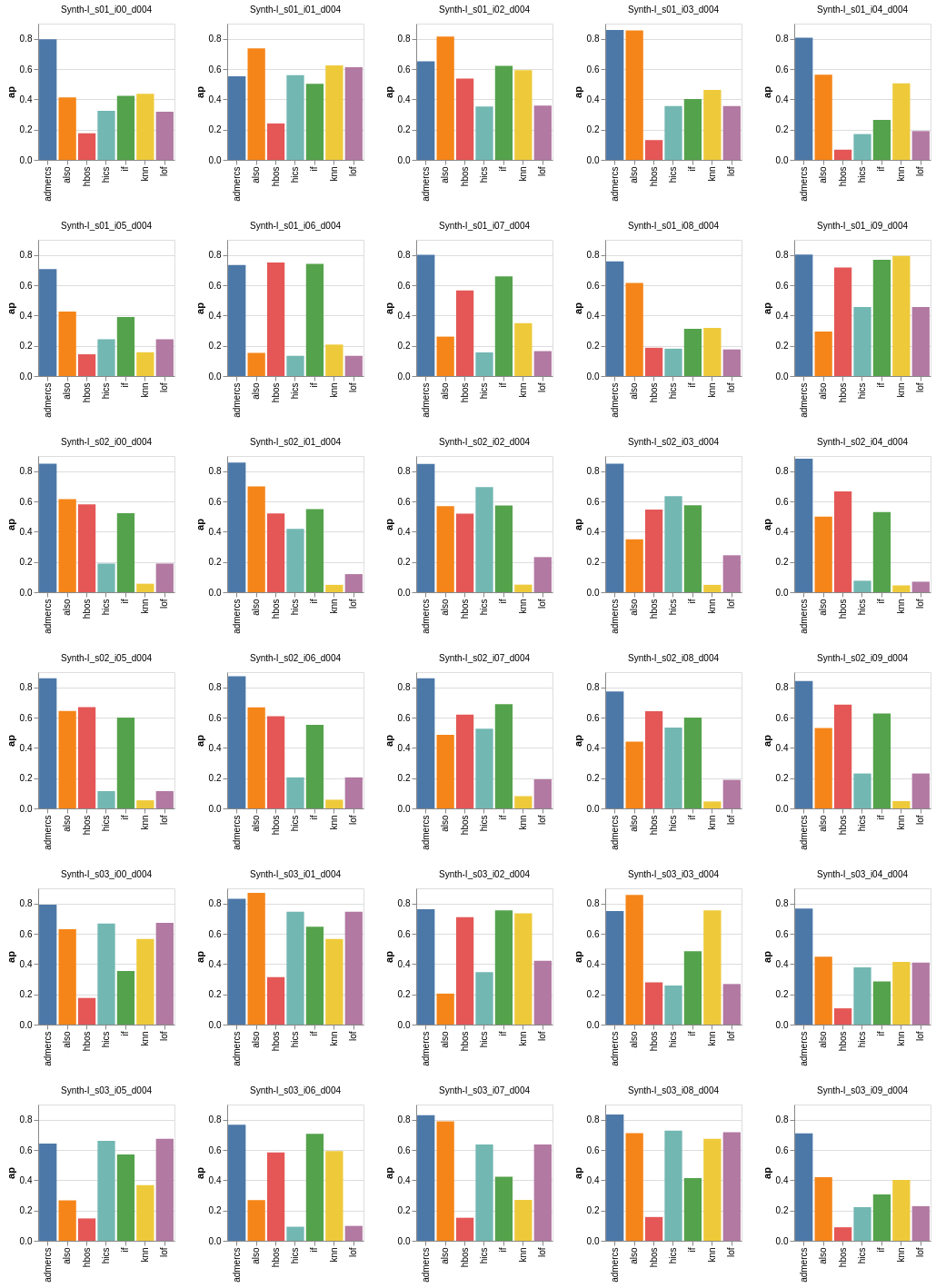}
    \caption{AP performance for each dataset and each algorithm for the Synth-I benchmark with a single parameter set for all datasets.}
    \label{fig:ap_per_dataset_anomcontext}
\end{figure}

The experimental results per dataset can be found in Fig. \ref{fig:auc_per_dataset_campos}- \ref{fig:ap_per_dataset_anomcontext}

\section{Parameter Sensitivity Plots for AD-MERCS}
\begin{figure}[h!]
    \centering
    \includegraphics[width = \textwidth]{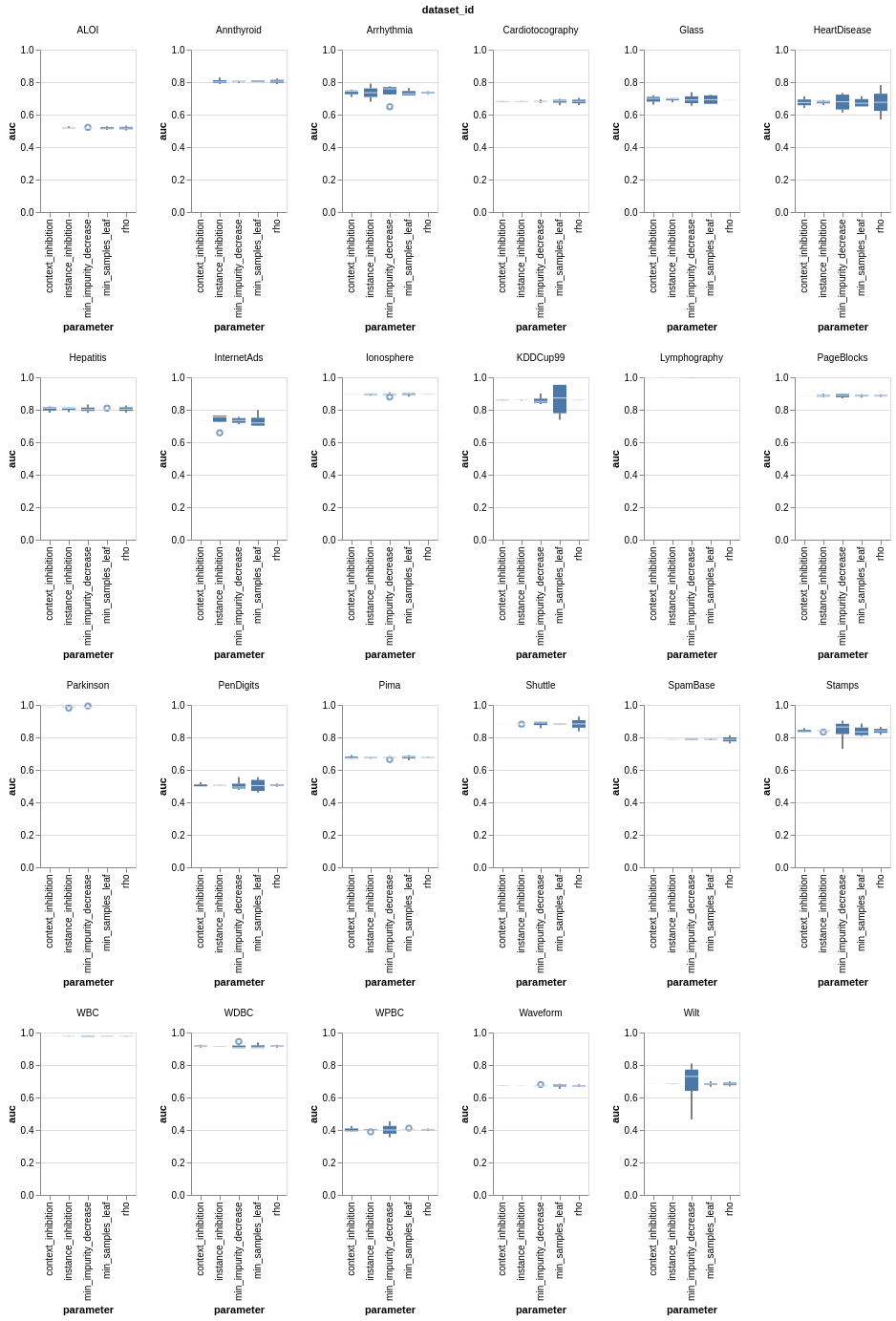}
    \caption{Parameter sensitivity plot for \admercs{} on the Campos benchmark per dataset}
    \label{fig:sensitivity_per_dataset_campos}
\end{figure}

\begin{figure}[h!]
    \centering
    \includegraphics[width = \textwidth]{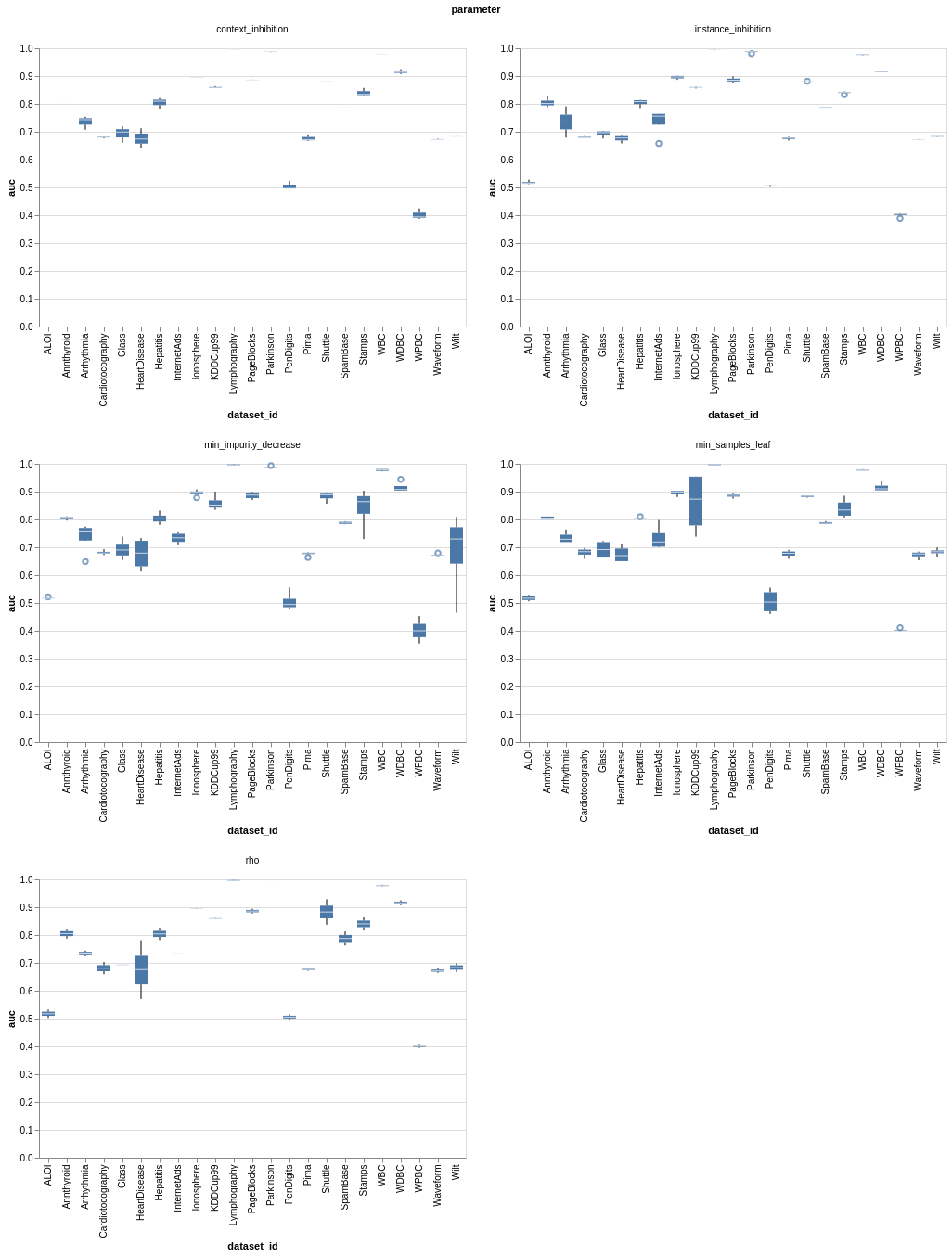}
    \caption{Parameter sensitivity plot for \admercs{} on the Campos benchmark per parameter}
    \label{fig:sensitivity_per_parameter_campos}
\end{figure}

\begin{figure}[h!]
    \centering
    \includegraphics[width = \textwidth]{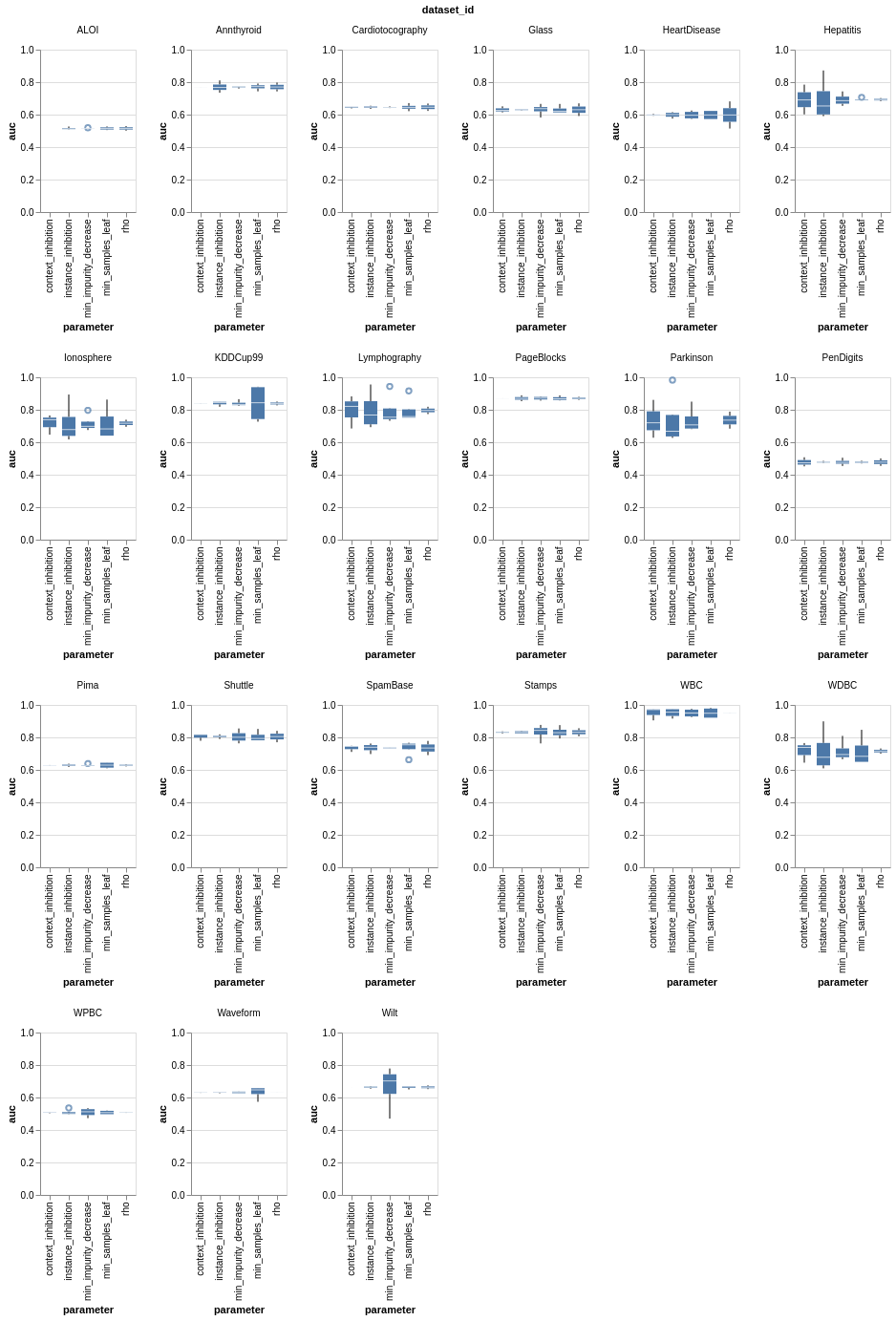}
    \caption{Parameter sensitivity plot for \admercs{} on the CamposHD benchmark per dataset}
    \label{fig:sensitivity_per_dataset_camposHD}
\end{figure}

\begin{figure}[h!]
    \centering
    \includegraphics[width = \textwidth]{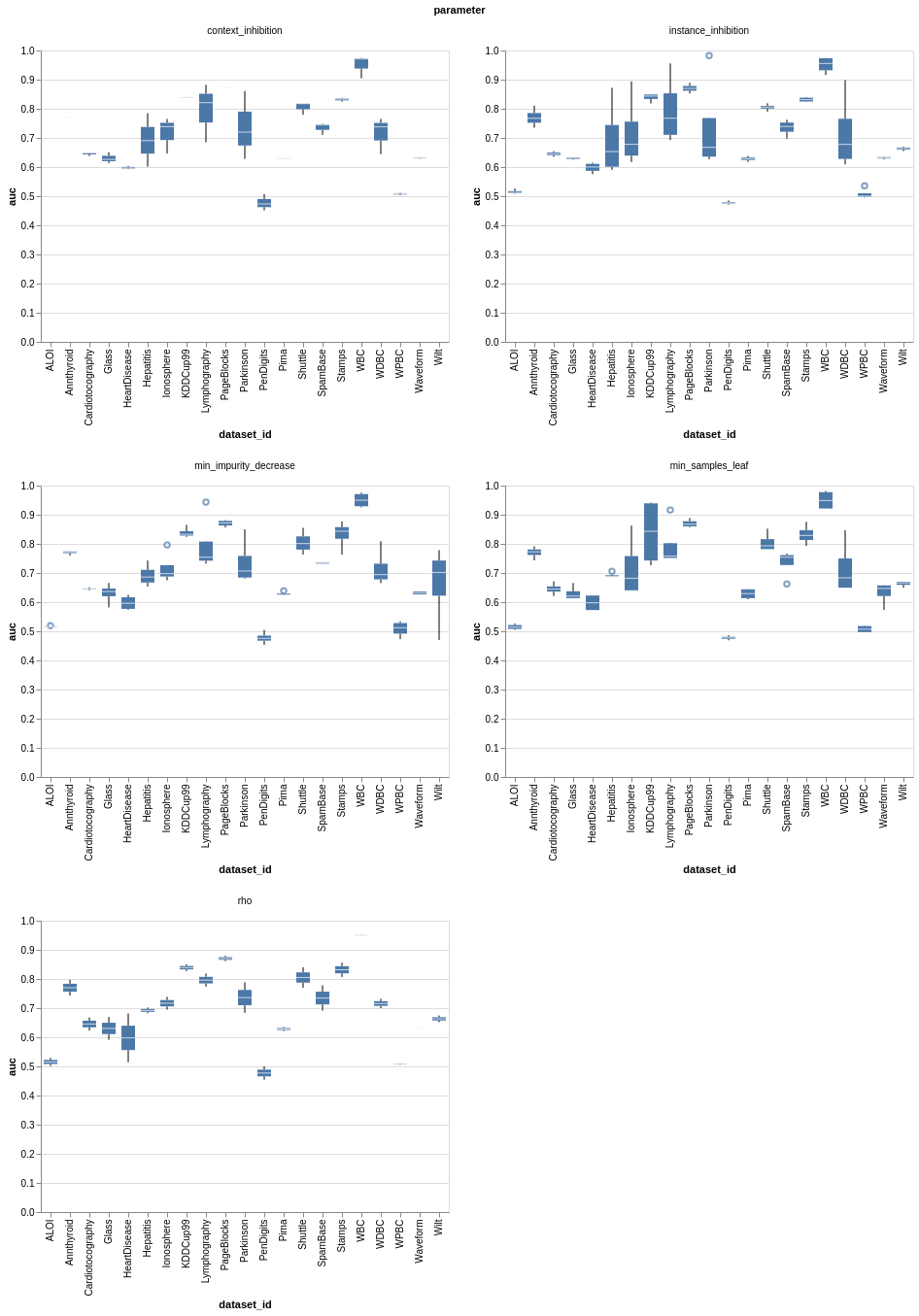}
    \caption{Parameter sensitivity plot for \admercs{} on the CamposHD benchmark per parameter}
    \label{fig:sensitivity_per_parameter_camposHD}
\end{figure}

\begin{figure}[h!]
    \centering
    \includegraphics[width = \textwidth]{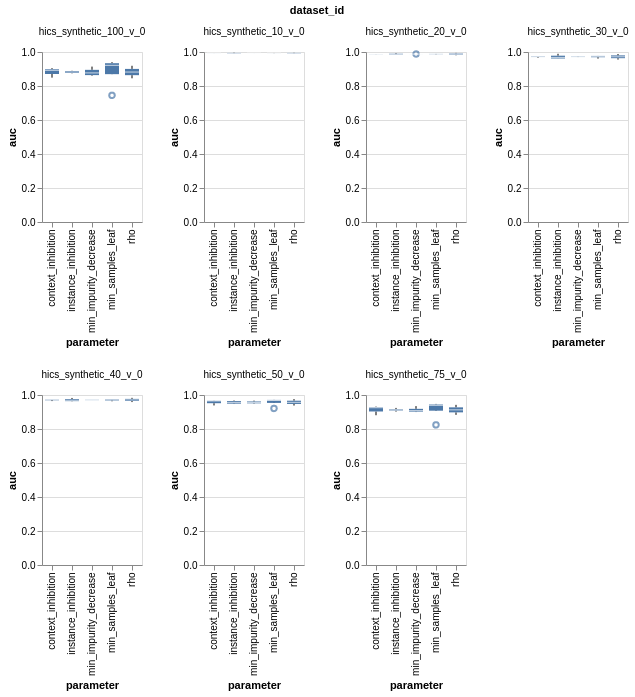}
    \caption{Parameter sensitivity plot for \admercs{} on the HiCS benchmark per dataset}
    \label{fig:sensitivity_per_dataset_hics}
\end{figure}

\begin{figure}[h!]
    \centering
    \includegraphics[width = \textwidth]{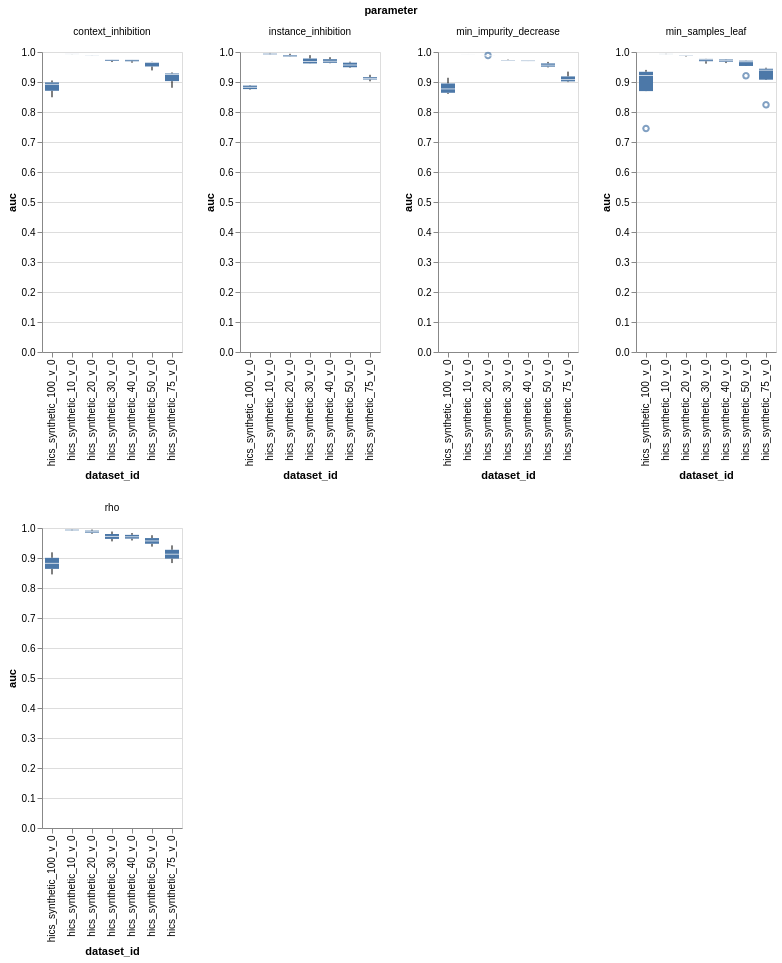}
    \caption{Parameter sensitivity plot for \admercs{} on the HiCS benchmark per parameter}
    \label{fig:sensitivity_per_parameter_hics}
\end{figure}

\begin{figure}[h!]
    \centering
    \includegraphics[width = \textwidth]{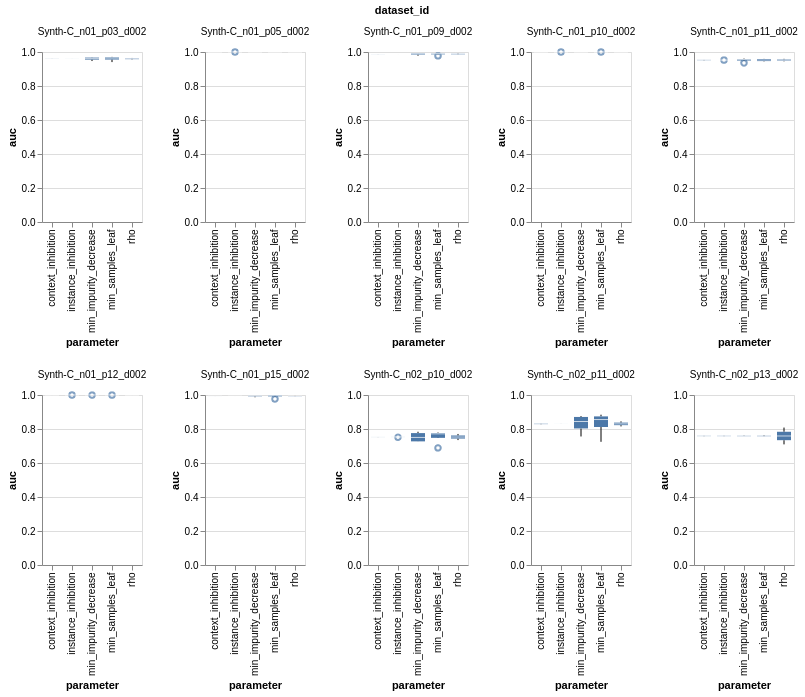}
    \caption{Parameter sensitivity plot for \admercs{} on the Synth-C benchmark per dataset}
    \label{fig:sensitivity_per_dataset_context}
\end{figure}

\begin{figure}[h!]
    \centering
    \includegraphics[width = \textwidth]{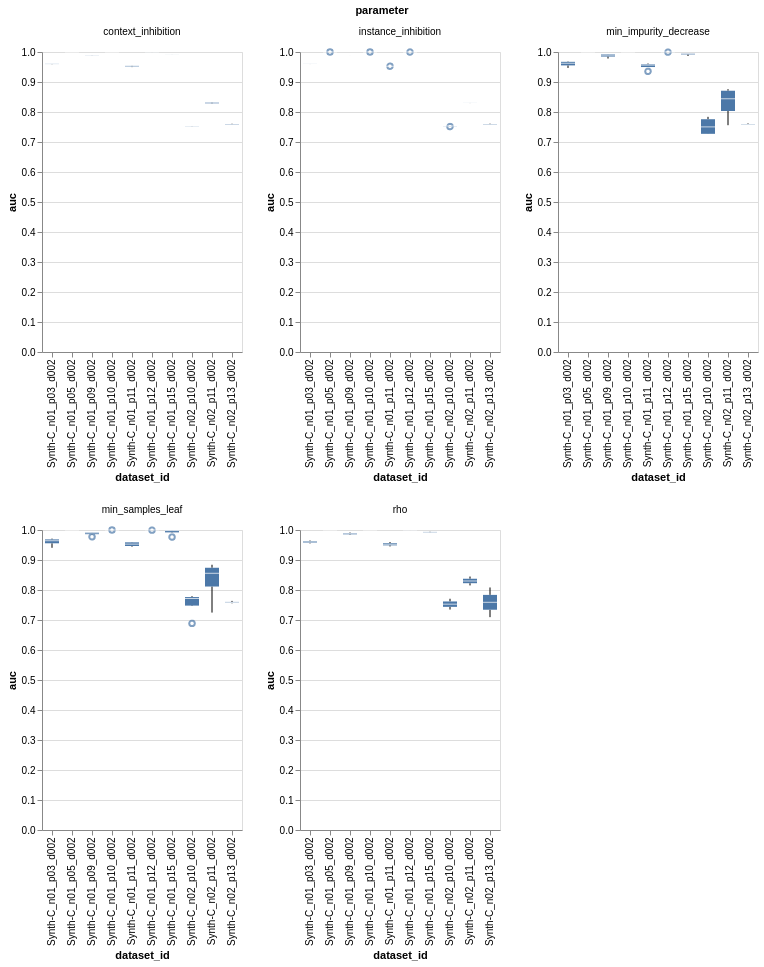}
    \caption{Parameter sensitivity plot for \admercs{} on the Synth-C benchmark per parameter}
    \label{fig:sensitivity_per_parameter_context}
\end{figure}

\begin{figure}[h!]
    \centering
    \includegraphics[width = \textwidth]{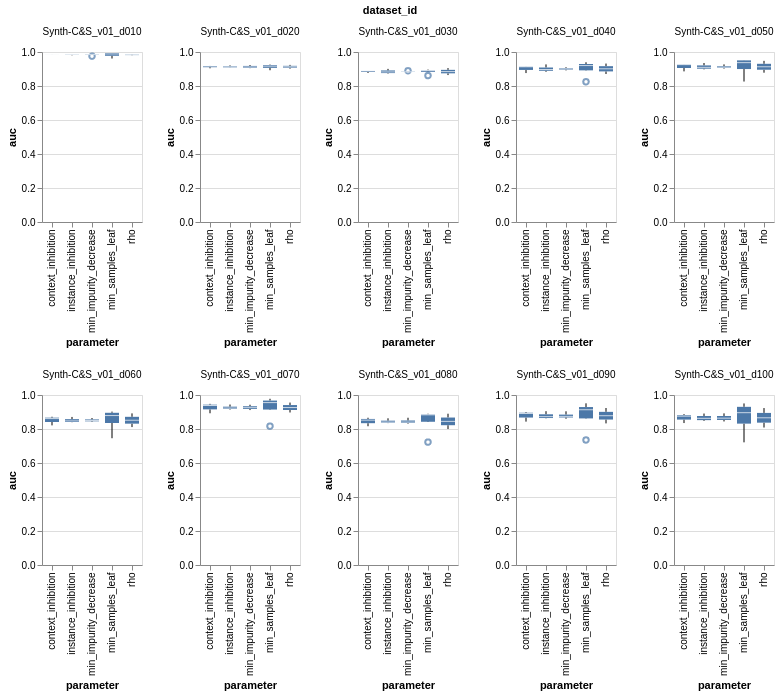}
    \caption{Parameter sensitivity plot for \admercs{} on the Synth-C\&S benchmark per dataset}
    \label{fig:sensitivity_per_dataset_context_subspace}
\end{figure}

\begin{figure}[h!]
    \centering
    \includegraphics[width = \textwidth]{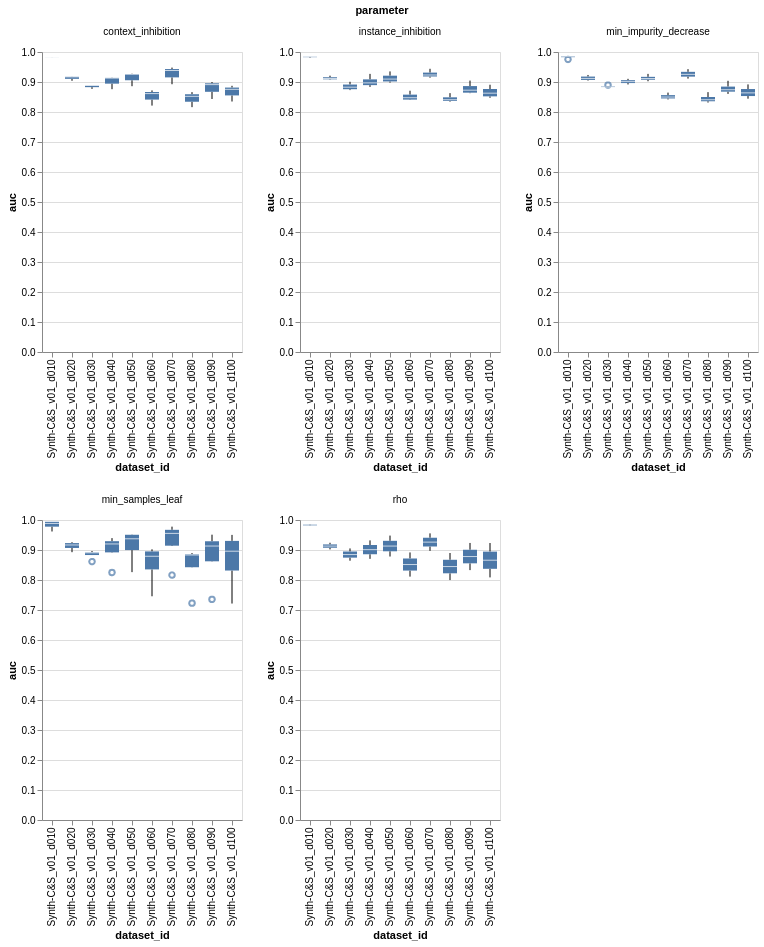}
    \caption{Parameter sensitivity plot for \admercs{} on the Synth-C\&S benchmark per parameter}
    \label{fig:sensitivity_per_parameter_context_subspace}
\end{figure}

\begin{figure}[h!]
    \centering
    \includegraphics[width = \textwidth]{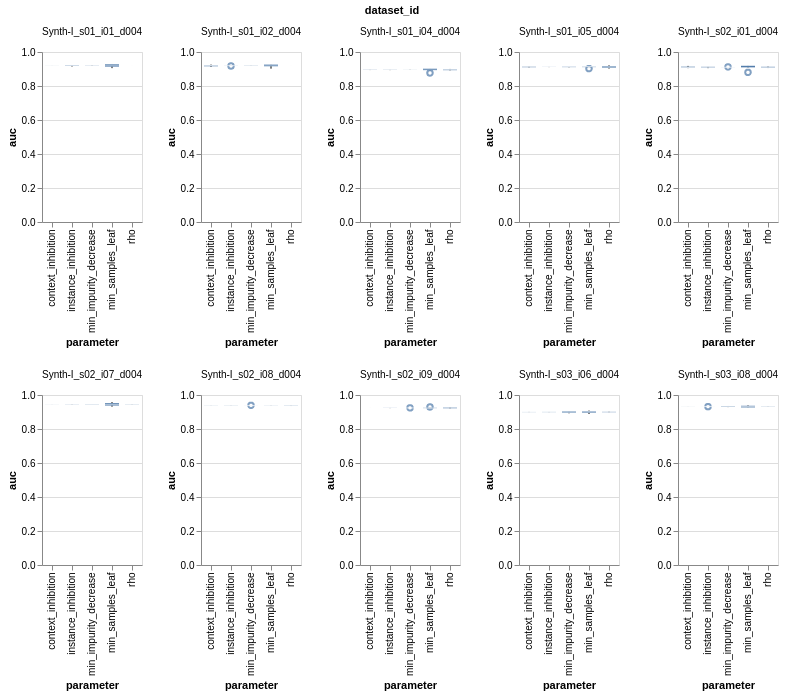}
    \caption{Parameter sensitivity plot for \admercs{} on the Synth-I benchmark per dataset}
    \label{fig:sensitivity_per_dataset_inlier}
\end{figure}

\begin{figure}[h!]
    \centering
    \includegraphics[width = \textwidth]{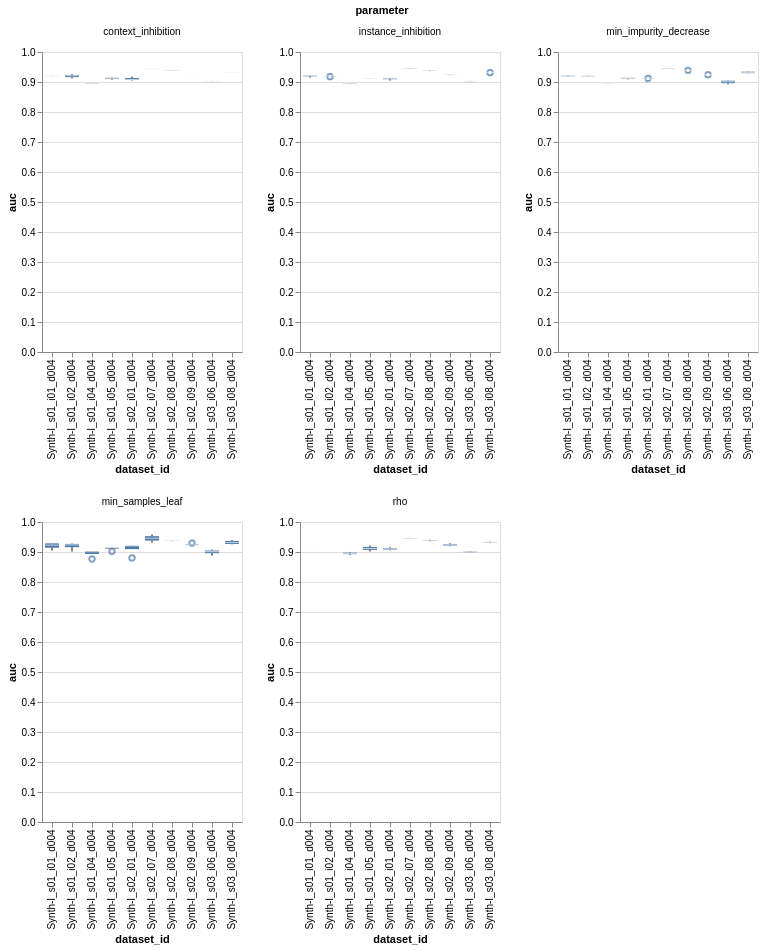}
    \caption{Parameter sensitivity plot for \admercs{} on the Synth-I benchmark per parameter}
    \label{fig:sensitivity_per_parameter_inlier}
\end{figure}

To give an indication how sensitive \admercs{} is to values of its parameters, we provide parameter sensitivity plots based on gridsearch results. 
We do so in the form of boxplots of the measured performances (in $AUC$) when varying one parameter and averaging out all the rest. 
For every benchmark, we provide two figures: one with a chart per dataset and one with a chart per parameter. 
These sensitivity plots can be found in Fig.~\ref{fig:sensitivity_per_dataset_campos}-\ref{fig:sensitivity_per_parameter_inlier}.

\begin{figure}[h!]
    \centering
    \includegraphics[width = 0.9\textwidth]{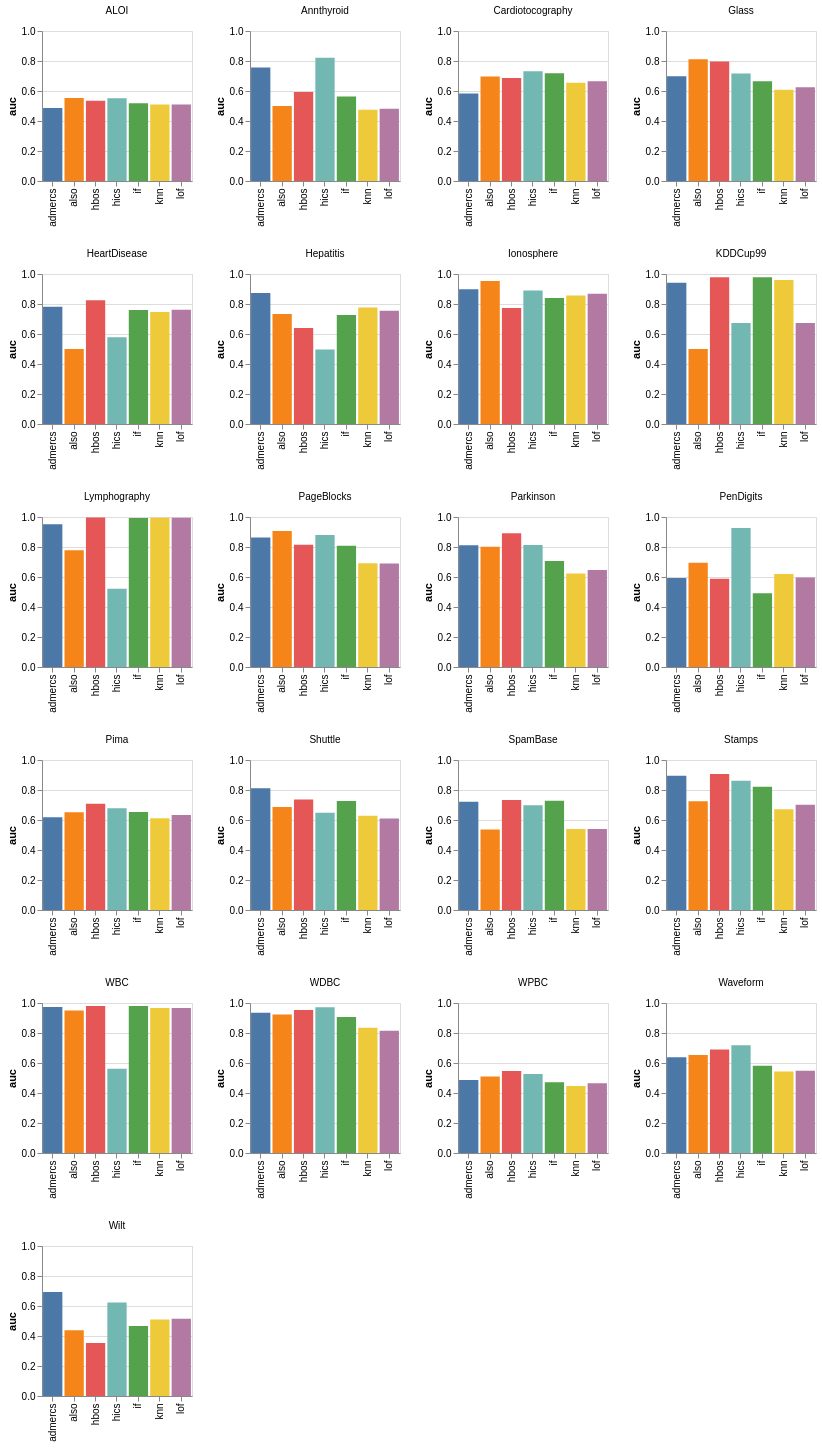}
    \caption{Individual AUC performance for each dataset and each algorithm (averaged over different subsample variations) for the CamposHD Benchmark with a single parameter set for all datasets.}
    \label{fig:auc_per_dataset_camposHD}
\end{figure}

\begin{figure}[h!]
    \centering
    \includegraphics[width = 0.9\textwidth]{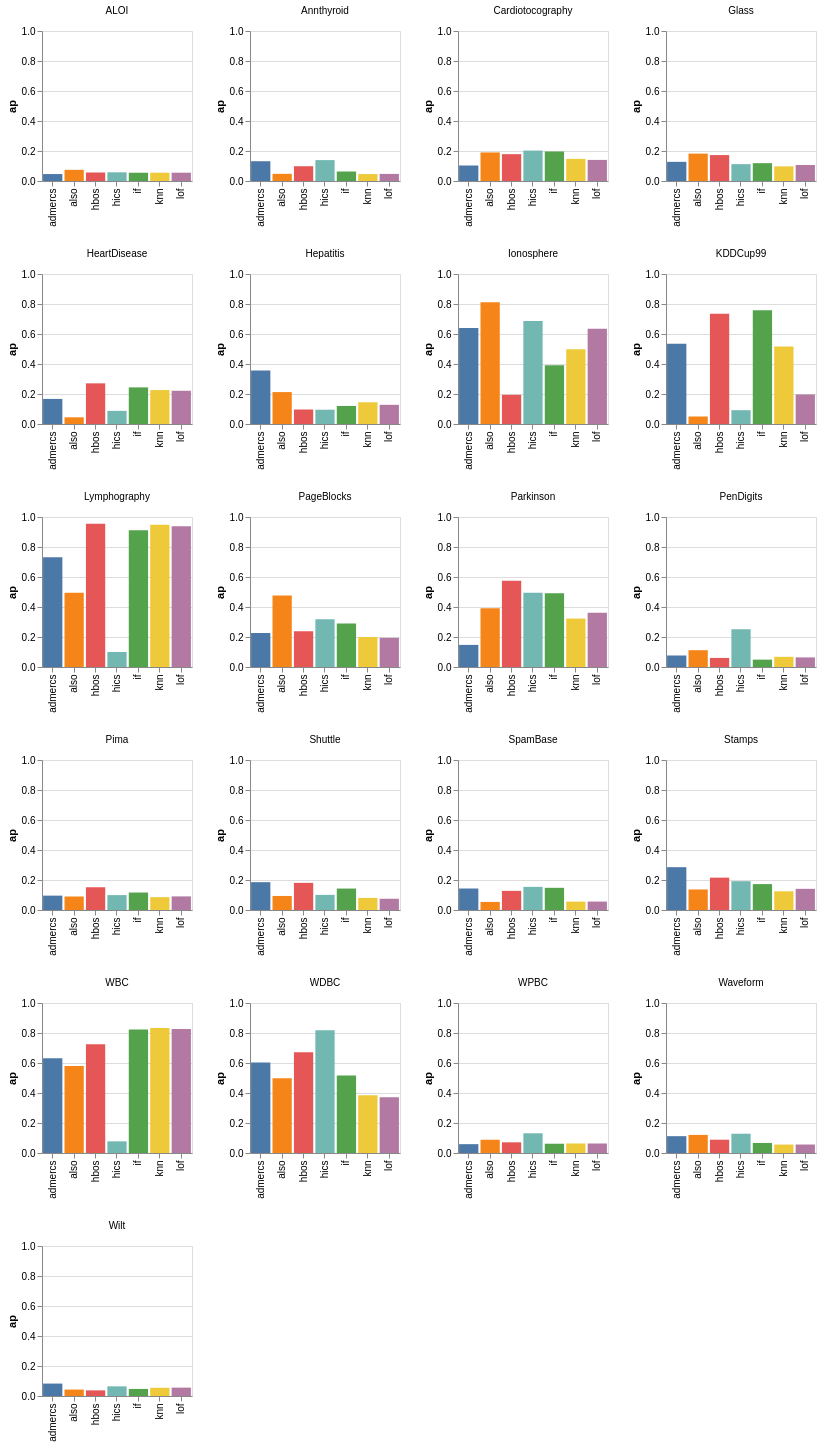}
    \caption{Individual AP performance for each dataset and each algorithm (averaged over different subsample variations) for the CamposHD benchmark with a single parameter set for all datasets.}
    \label{fig:ap_per_dataset_camposHD}
\end{figure}

\begin{figure}[h!]
    \centering
    \includegraphics[width = 0.9\textwidth]{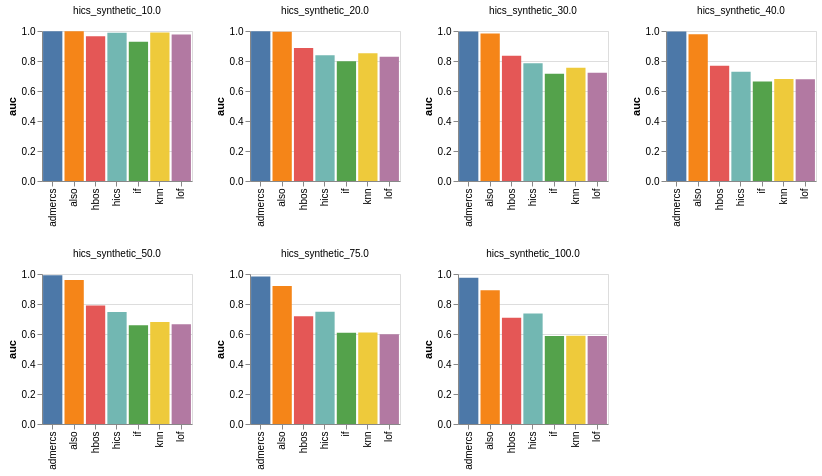}
    \caption{Individual AUC performance for each dataset and each algorithm (averaged over different versions to be able to see performance decrease due to the increase in dimensionality clearly) for the HiCS Benchmark with a single parameter set for all datasets.}
    \label{fig:auc_per_dataset_camposHD}
\end{figure}

\begin{figure}[h!]
    \centering
    \includegraphics[width = 0.9\textwidth]{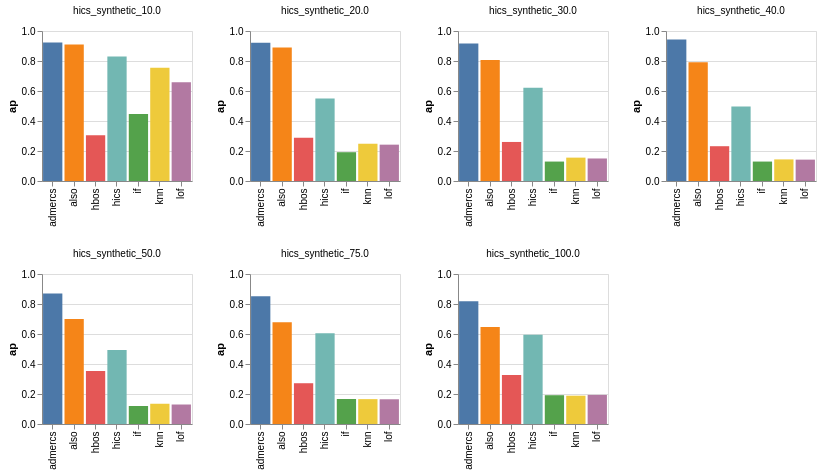}
    \caption{Individual AP performance for each dataset and each algorithm (averaged over different versions to be able to see the performance decrease due to the increase in dimensionality clearly) for the HiCS benchmark with a single parameter set for all datasets.}
    \label{fig:ap_per_dataset_camposHD}
\end{figure}

